\documentclass{article}

\usepackage[preprint]{neurips_2026}    % arXiv preprint

\usepackage[utf8]{inputenc}
\usepackage[T1]{fontenc}

\usepackage{hyperref}
\usepackage{url}
\usepackage{microtype}
\usepackage{xspace}
\usepackage{xcolor}

\usepackage{amsfonts}
\usepackage{amssymb}
\usepackage{mathtools}
\usepackage{amsthm}
\usepackage{nicefrac}
\usepackage{bm}

\usepackage{booktabs}
\usepackage{multirow}
\usepackage{makecell}

\usepackage{graphicx}
\usepackage{float}
\usepackage{caption}
\usepackage{subcaption}

\usepackage{cleveref}
\crefname{equation}{Eq.}{Eqs.}
\crefname{table}{Tab.}{Tabs.}
\crefname{section}{Sec.}{Secs.}
\crefname{figure}{Fig.}{Figs.}

\usepackage{algorithm}
\usepackage{algpseudocode}

\usepackage{enumitem}
\usepackage{colortbl}
\definecolor{burntorange}{RGB}{204, 85, 0}      % Burnt Orange
\definecolor{crimson}{RGB}{220, 20, 60}        % Crimson
\definecolor{teal}{RGB}{0, 128, 128}           % Teal
\definecolor{indigo}{RGB}{75, 0, 130}          % Indigo
\definecolor{royalblue}{RGB}{65, 105, 225}     % Royal Blue
\definecolor{magenta}{RGB}{255, 0, 255}        % Magenta

\definecolor{darkgreen}{rgb}{0,0.6,0}

\newtheorem*{hypothesis}{Hypothesis}

\def\eqref#1{equation~\ref{#1}}

\def\1{\bm{1}}

\def\vv{{\bm{v}}}

\def\vz{{\bm{z}}}

\colorlet{yellow}{green!10!orange}

\definecolor{darkgreen}{rgb}{0.0, 0.5, 0.0}

\title{ArtCrafter: Feed-Forward Generation of Articulated 3D Object with Analytic Joint Derivation}

\author{ \begin{tabular}{ccc} Minh Canh Tu$^{*}$ & Quang-Binh Nguyen$^{*\dagger}$ & Khoi Nguyen \end{tabular}\\[0.5em] Qualcomm AI Research$^{\ddagger}$ }

\definecolor{mydarkblue}{rgb}{0,0.08,1}
\definecolor{mydarkgreen}{rgb}{0.02,0.6,0.02}
\definecolor{myred}{rgb}{1.0,0.0,0.0}
\definecolor{myred2}{rgb}{0.7,0.1,0.1}
\definecolor{mydarkblue2}{rgb}{0.05,0.1,0.7}
\definecolor{mypurple}{rgb}{111,0,255}
\definecolor{mypurple2}{rgb}{111,0,111}

\newcommand{\myheading}[1]{\vspace{2mm}\noindent{\textbf{#1}}}

\usepackage{pifont}

\begin{document}

\maketitle

\begingroup \renewcommand{\thefootnote}{\fnsymbol{footnote}} \footnotetext[1]{Equal contribution.} \footnotetext[2]{Work done while at Qualcomm.} \footnotetext[3]{Qualcomm AI Research is an initiative of Qualcomm Technologies, Inc.} \endgroup

\begin{abstract} Articulated 3D assets are essential for robotic simulation and training, yet their generation from casual images poses a key challenge: recovering part geometry and kinematics that remain mutually consistent. We present ArtCrafter, a novel feed-forward generative model for controllable generation of articulated 3D objects from a pair of images, one depicting the object in the rest state and the other in the open state. At the core of ArtCrafter is a DiT-based generative model that jointly produces part-level meshes for both articulation states in a single forward pass. Inspired by Slot Attention, its denoising process operates on a $1+2N$ slot layout comprising one shared base slot and $N$ paired rest/open slots, encouraging each pair to bind to the same kinematic part. To ensure geometrically consistent and disjoint part generation, we introduce Triplet Attention to enforce cross-state geometric consistency between paired rest/open slots anchored through the shared base slot, and a repulsion loss $\mathcal{L}_{\mathrm{rep}}$ on decoded part point clouds to prevent inter-part penetration throughout denoising. Given the resulting paired meshes, URDF joint parameters--- type, axis, origin, and range---are derived analytically from the rigid transformation between each rest/open pair, providing an interpretable alternative to learned joint-parameter regression. Extensive experiments on both in-distribution and out-of-distribution datasets demonstrate that ArtCrafter outperforms existing method under the same dual-state setting in geometry quality and articulation precision, while remaining competitive with methods operating under different input settings and paradigms. \end{abstract}

\section{Introduction}
\label{sec:intro}

Articulated 3D assets are fundamental to robotic simulation and embodied AI, where agents interact with objects composed of movable parts. Unlike static 3D models, simulation-ready articulated assets must represent both part-level geometry and the kinematic constraints governing how parts move. However, manually constructing such assets requires substantial modeling effort and expert annotation, making it difficult to achieve the scale and diversity needed for general-purpose robot training. PartNet-Mobility~\cite{partnetmobilitydataset}, a widely used articulated-object dataset, contains only approximately 2.3K manually annotated objects, underscoring the need for scalable automated asset creation.

Automatically generating articulated assets from images is fundamentally more challenging than ordinary 3D generation. Beyond reconstructing appearance and geometry, a method must decompose the object into kinematic parts and recover joint parameters that remain geometrically consistent with the generated meshes. Current 3D generators either produce whole-object meshes with no internal structure~\cite{triposg,trellis,clay,hunyuan3d}, or decompose objects into parts defined primarily by geometric saliency without modeling their kinematic function~\cite{partcrafter,holopart,partgen}. Neither class directly produces articulated assets suitable for physics simulation.

\begin{table}[t]
\centering
\caption{%
  \textbf{Taxonomy of articulated 3D methods by input regime and paradigm.}
  ArtCrafter is the only generative method conditioned on two articulation states.
}
\label{tab:taxonomy}
\footnotesize
% \small
\setlength{\tabcolsep}{4pt}
\renewcommand{\arraystretch}{1.15}
\begin{tabular*}{\textwidth}{@{\extracolsep{\fill}}p{2cm}p{4.5cm}p{2.0cm}p{4.2cm}}
\toprule
\textbf{Paradigm} & \textbf{Single-state} & \textbf{Dual-state} 
& \textbf{Multi-view / multi-state} \\
\midrule
\textbf{Retrieval}
  % & \xmark
  & Articulate-Anything~\cite{liu2024articulateanything},
  SINGAPO~\cite{liu2024singapo},
    URDFormer~\cite{urdfformer}
  & DIPO~\cite{dipo}
  & ---
\\
\addlinespace
\textbf{Segmentation}
  % & \xmark
  & URDF-Anything~\cite{urdfanything}
  & ---
  & URDF-Anything~\cite{urdfanything}
\\
\addlinespace
\textbf{Optimization}
  % & \xmark
  & ---
  & ---
  & PARIS~\cite{liu2023paris},
    ArtGS~\cite{ye2024artgs},
    LARM~\cite{larm},
    FreeArt3D~\cite{freeart3d},
    REArtGS~\cite{reartgs}
\\
\addlinespace
\textbf{Generation}
  % & \cmark
  & PAct~\cite{mu2024pact},
    SPARK~\cite{spark},
    URDF-Anything+~\cite{urdfanythingplus},
    PhysX-Anything~\cite{physxanything}
  & \textbf{ArtCrafter (ours)}
  & ART~\cite{liu2024art}
\\
\bottomrule
\end{tabular*}
\vspace{-10pt}
\end{table}

Existing articulated object creation methods differ along two axes: input observation regime and geometry production mechanism, as summarized in \cref{tab:taxonomy}. For input, \textit{multi-view/multi-state} methods~\cite{liu2023paris, ye2024artgs, liu2024art, larm, freeart3d} rely on multiple viewpoints, multiple articulation states, or both to provide sufficient geometric and motion observations for reconstruction. While these methods can reconstruct articulated objects with high accuracy, they require extensive data acquisition effort that is hard to scale. \textit{Single-state} or \textit{single-image} methods utilize minimal input but lack explicit articulation information, so they must predict kinematic behavior based purely on their semantic understanding of the object and its parts. DIPO~\cite{dipo} introduced the \textit{dual-state} input setting -- a rest and open image pair -- that provides essential motion cues and connectivity information without dense input requirement. This input setting provides a practical middle ground between single-state and multi-view/multi-state approaches, reducing ambiguity in part motion and spatial relationships while avoiding the dense data acquisition required by the latter. For geometry, \textit{retrieval-based}~\cite{liu2024singapo, urdfformer, dipo, liu2024articulateanything, li2024cage, lei2023nap}, \textit{segmentation-based}~\cite{particulate, urdfanything, geopard, motionanymesh, fan2021ditto}, \textit{optimization-based}~\cite{liu2023paris,ye2024artgs, larm, freeart3d}, and \textit{generative}~\cite{spark, mu2024pact, monoart, liu2024art, urdfanythingplus, physxanything} methods differ in whether they retrieve, decompose, fit, or synthesize part geometries, respectively. We adopt the \textbf{dual-state and generative paradigm}, combining direct kinematic evidence with unconstrained geometry synthesis.

In this paper, we present \textbf{ArtCrafter}, a two-stage feed-forward framework for controllable generation of articulated 3D objects from a rest/open image pair. ArtCrafter leverages the semantic priors of a pretrained TripoSG~\cite{triposg} DiT to synthesize geometry from only one image per state, while retaining the direct motion evidence unavailable to single-image generation. In the first stage, the DiT jointly generates part-level meshes for both articulation states through a \textbf{$1+2N$ slot layout}, comprising one shared base slot and $N$ paired rest/open slots. Inspired by Slot Attention~\cite{locatello2020slotattention}, jointly explaining both states encourages each slot pair to bind to the same kinematic part. We further introduce \textbf{triplet attention} to couple each rest/open pair through the shared base slot and enforce cross-state geometric consistency, together with a repulsion loss $\mathcal{L}_{\mathrm{rep}}$ on decoded part point clouds to prevent inter-part penetration throughout denoising. In the second stage, joint parameters--- type, axis, origin, and range---are derived analytically from the rigid transformation between each generated rest/open part pair, grounding kinematics directly in the generated geometry without learned joint-parameter regression or per-instance optimization. Extensive experiments on both in-distribution and out-of-distribution benchmarks demonstrate improvements under the directly comparable dual-state setting, while showing competitive performance against methods operating under different input settings and paradigms.

In summary, our main contributions are as follows:

\begin{enumerate}[noitemsep, topsep=0pt] 
    \item \textbf{Dual-state generative formulation with analytic articulation recovery.} We generate paired part geometries across two articulation states and derive joint parameters analytically from their rigid transformations, grounding kinematics directly in geometry.
    \item \textbf{$1+2N$ slot DiT with triplet attention and $\mathcal{L}_{\mathrm{rep}}$.} A structured slot layout binds rest/open representations to the same kinematic parts; triplet attention enforces cross-state consistency, while $\mathcal{L}_{\mathrm{rep}}$ promotes inter-part disjointness.
    \item \textbf{Extensive evaluation.} Experiments on both in-distribution and out-of-distribution benchmarks demonstrate significant improvements over directly comparable two-state baselines in geometry quality and articulation precision, while showing competitive performance against methods operating under different input settings and paradigms.
\end{enumerate}

\section{Related Work}
\label{sec:related}

\textbf{3D object mesh generation.} 3D-native generative methods learn shape distributions over compact 3D latent representations. Two representation families are prominent. \textbf{Vecset-based methods}~\cite{zhang2023shape2vecset, clay, triposg, hunyuan3d} represent shapes as unordered sets of latent vectors and enable efficient Transformer-based 3D generation. \textbf{Sparse voxel-based methods}~\cite{trellis, ren2024xcube, he2025triposf, wu2025direct3ds2, li2025sparc3d} instead anchor latent features to sparse spatial locations, preserving explicit spatial structure for high-fidelity geometry generation. ArtCrafter adopts the vecset representation of TripoSG~\cite{triposg}, whose flexible set-based structure naturally accommodates multiple part-level latent slots without imposing a shared spatial grid. 

\textbf{Articulated object creation} methods differ along two axes: input observation regime and geometry production mechanism, as summarized in \cref{tab:taxonomy}. For geometry production, retrieval-based methods construct part geometries by retrieving template meshes from a fixed database. Non-retrieval methods obtain part geometries by decomposing a reconstructed monolithic shape into parts (\textit{segmentation-based}), iteratively fitting object-specific geometric representations to observations (\textit{optimization-based}), or directly synthesizing part meshes with learned generative models (\textit{generative}).

\textbf{Single-state methods.} Single-state methods create articulated 3D assets from one image, with recent approaches predominantly adopting generative models~\cite{mu2024pact, spark, urdfanythingplus, physxanything}. PAct~\cite{mu2024pact} and SPARK~\cite{spark} generate part-level geometry with part-aware diffusion models, URDF-Anything+~\cite{urdfanythingplus} autoregressively generates parts and joints, and PhysX-Anything~\cite{physxanything} predicts a physical representation decoded into part geometry and articulation. SINGAPO~\cite{liu2024singapo} instead generates abstract part structure and articulation before retrieving concrete meshes, while URDFormer~\cite{urdfformer} and Articulate-Anything~\cite{liu2024articulateanything} retrieve and assemble geometry based on predicted object structure. Following a segmentation-based paradigm, URDF-Anything~\cite{urdfanything} reconstructs a monolithic 3D point cloud and jointly segments its parts and predicts kinematic parameters with a 3D MLLM. Without observed motion, however, single-state methods must infer ambiguous part structure and kinematics from appearance alone.

\textbf{Multi-view or multi-state methods.}
Under richer observation settings, URDF-Anything~\cite{urdfanything} additionally supports multi-view input by reconstructing its point-cloud representation from multiple images. ART~\cite{liu2024art} instead adopts a feed-forward approach that directly reconstructs part-level geometry and predicts articulation parameters from sparse multi-state images. Optimization-based methods~\cite{liu2023paris, ye2024artgs, freeart3d, reartgs} instead fit object-specific geometric representations, such as neural implicit fields, 3D Gaussians, or occupancy fields, together with articulation parameters to the input observations through per-instance optimization. While capable of detailed reconstruction, their iterative inference and richer observation requirements limit scalability.

\textbf{Dual-state methods.} The dual-state setting uses one image from each of two articulation states, providing direct visual evidence of part motion while requiring substantially fewer observations than multi-view/multi-state reconstruction. DIPO~\cite{dipo} established this dual image-based setting, conditioning a diffusion transformer on the rest/open image pair to predict part layouts and joint parameters, and concrete part meshes are subsequently retrieved from a fixed database. Ditto~\cite{fan2021ditto} similarly exploits motion evidence from two states, but reconstructs geometry and articulation from paired partial 3D point clouds rather than images, making its input setting distinct from dual-image methods. Building on the dual-image setting, ArtCrafter directly synthesizes geometrically consistent rest/open part meshes with a generative DiT and analytically derives joint parameters from their rigid transformations, avoiding DIPO~\cite{dipo} mesh-retrieval step and grounding the inferred kinematics in geometry.
\section{Method}
\label{sec:method}

\begin{figure}[t]
  \centering
  \includegraphics[width=.85\linewidth]{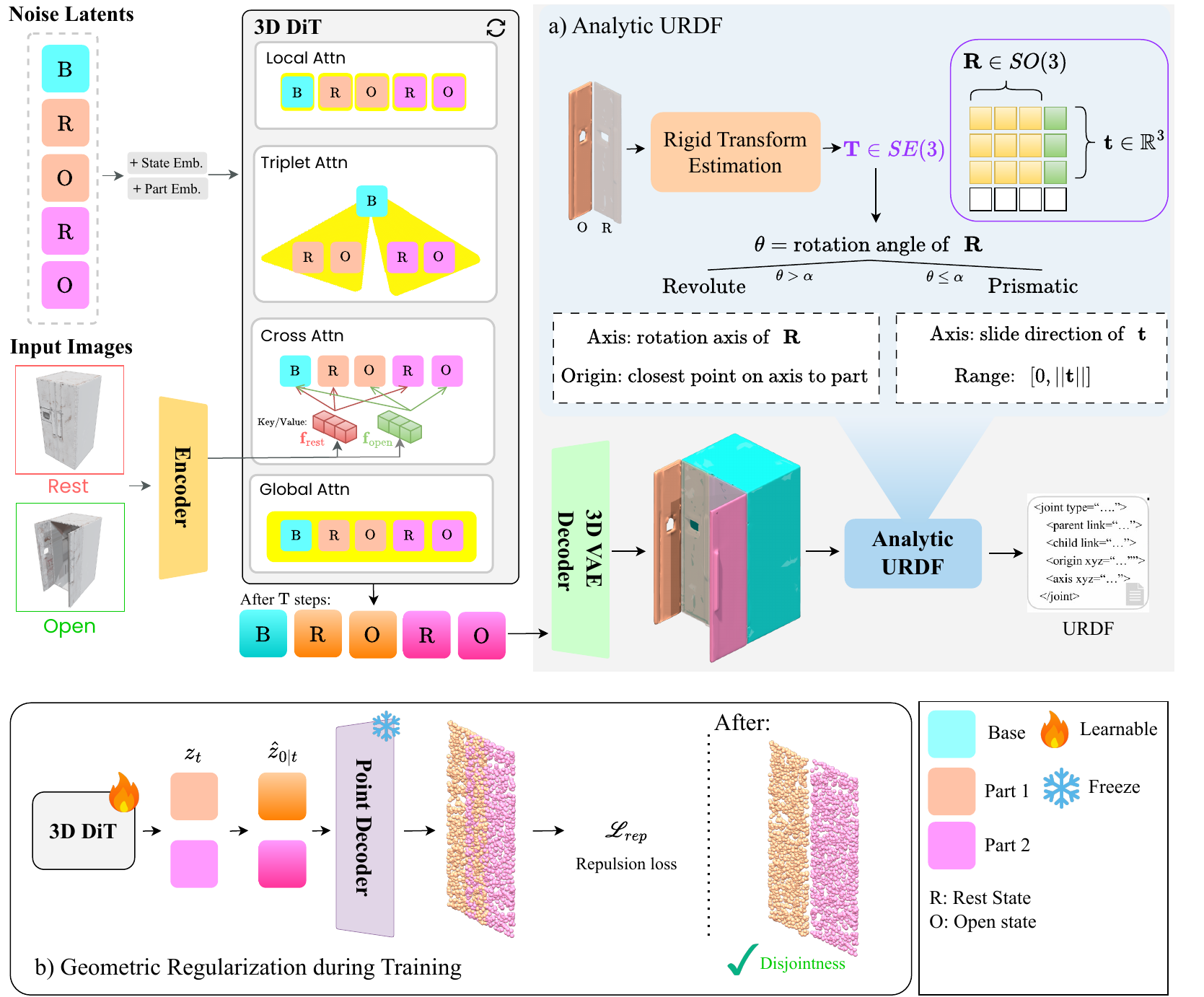}
  \caption{%
    \textbf{ArtCrafter pipeline.}
    Two input images ($\mathbf{I}_{\mathrm{rest}}$, $\mathbf{I}_{\mathrm{open}}$) are encoded into state features conditioning $2N+1$ noisy latent slots — base $B$, rest slots $R^1\!\ldots\!R^N$, open slots $O^1\!\ldots\!O^N$ — each augmented with learnable part and state embeddings. Each \textbf{3D DiT} block refines slots through: \emph{Local Attn} (intra-slot), \emph{Triplet Attn} ($R^i \leftrightarrow O^i$ anchored through $B$), \emph{State Cross Attn} (rest$\to\mathbf{f}_{\mathrm{rest}}$, open$\to\mathbf{f}_{\mathrm{open}}$, base$\to$both), and \emph{Global Attn} (all $2N+1$ slots). The frozen VAE decoder produces part meshes from denoised slots.
    \emph{(a) Analytic joint derivation:} joint parameters are derived deterministically from $\mathbf{T}_i = \mathbf{T}_{\mathrm{open},i} \cdot \mathbf{T}_{\mathrm{rest},i}^{-1}$ with no learned head.
    \emph{(b) Geometric regularization}: the frozen point decoder decodes $\hat{\mathbf{z}}_{0|t}$ to per-slot point clouds; $\mathcal{L}_{\mathrm{rep}}$ penalizes inter-part overlap.
    % \minh{Move GlobalAttn to top, rename Analytic URDF with Analytic Joint Derivation}
  }
  \label{fig:pipeline}
  \vspace{-10pt}
\end{figure}

\begin{figure}[t]
  \centering
  \includegraphics[width=.9\linewidth]{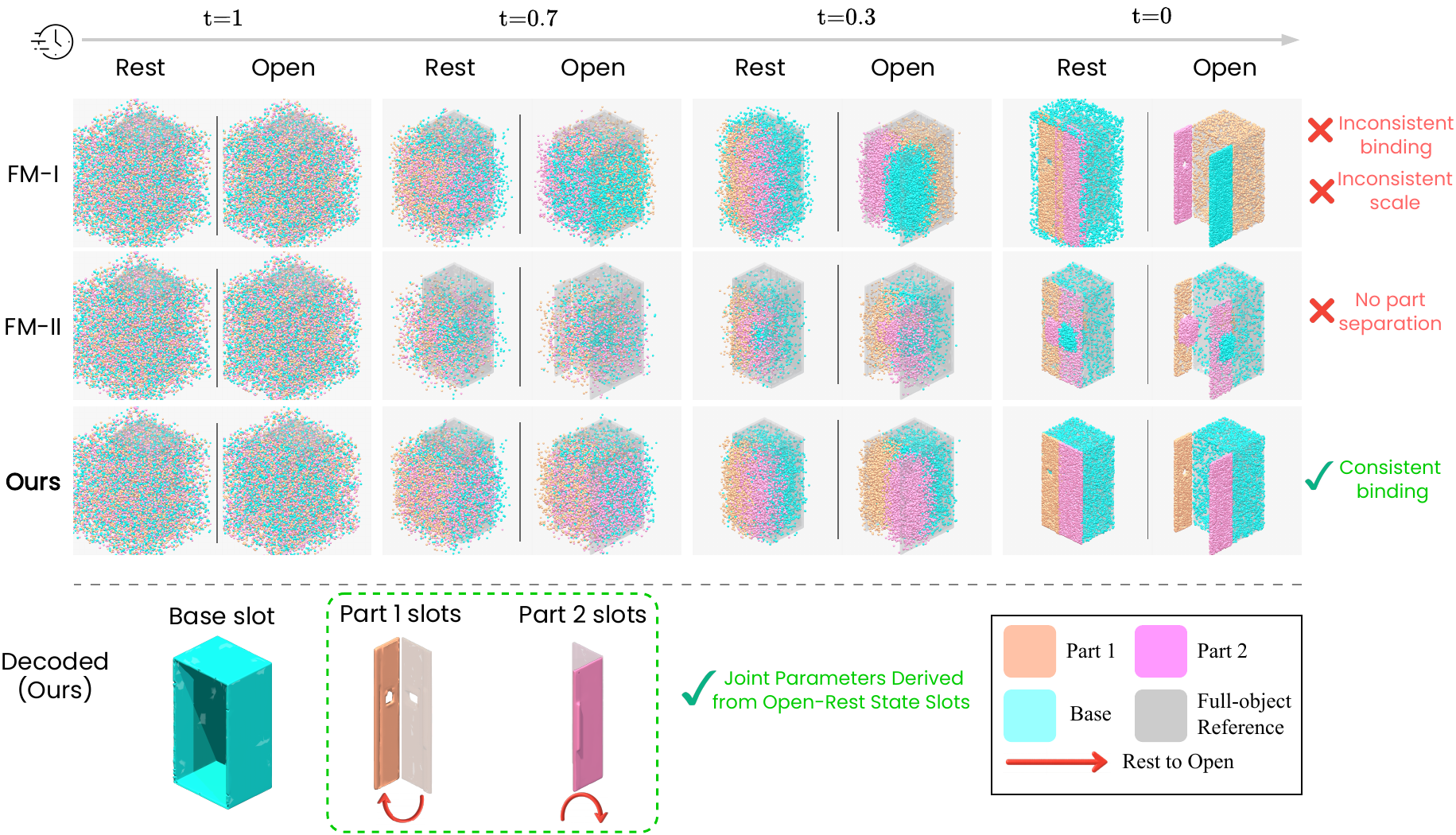}
  \caption{%
    \textbf{Per-slot decoded geometry across denoising timesteps.}
    Rows compare ArtCrafter (Ours), FM-I (independent per-state), and FM-II (unstructured joint generation) on the same object; columns progress from pure noise ($t=1$) to clean output ($t=0$). Each cell shows a split render of rest (left) and open (right) states overlaid on the object silhouette (gray); colors encode part identity: base (cyan), part~1 (orange), part~2 (pink). ArtCrafter achieves consistent cross-state binding by $t=0$; FM-I produces inconsistent binding and incompatible part scale across the two independent runs; FM-II shows no clean part separation in either state. \emph{Bottom:} decoded meshes per slot at $t=0$, from which joint parameters are derived analytically. 
    % See \cref{sec:slots} for design implications.%
}
  \label{fig:slot_trajectory}
  \vspace{-10pt}
\end{figure}

\subsection{Problem Statement}
\label{sec:problem}

Given two RGB images $\mathbf{I}_{\mathrm{rest}}, \mathbf{I}_{\mathrm{open}} \in \mathbb{R}^{H \times W \times 3}$ depicting an articulated object in its rest and fully open states, and the number of movable parts $N$, either specified by the user or estimated by a VLM from the two input images, ArtCrafter generates $1+2N$ part-level meshes: a shared static base mesh $\mathcal{M}_{\mathrm{base}}$ and $N$ pairs of movable-part meshes $\{(\mathcal{M}_{\mathrm{rest}}^i, \mathcal{M}_{\mathrm{open}}^i)\}_{i=1}^{N}$. The corresponding joint parameters $\{(\mathcal{J}_T^i, \mathcal{J}_A^i, \mathcal{J}_O^i, \mathcal{J}_L^i)\}_{i=1}^{N}$---type, axis, origin, and range---are then derived analytically from the rigid transformation between each rest/open mesh pair and assembled with the generated meshes into a physics-executable URDF.

\subsection{Overview}
\label{sec:overview}

ArtCrafter builds on TripoSG~\cite{triposg}, a large-scale 3D generation DiT pretrained on Objaverse~\cite{deitke2023objaverse}. Its VAE encodes each part point cloud into vecset latents $\vz \in \mathbb{R}^{L \times d}$ with $L$ tokens and decodes them into meshes via marching cubes. As illustrated in \cref{fig:pipeline}, a vision encoder extracts state features $\mathbf{f}_{\mathrm{rest}}$ and $\mathbf{f}_{\mathrm{open}}$ that condition $1+2N$ noisy latent slots---one base and $N$ rest/open pairs---jointly denoised by the DiT. The DiT backbone alternates between local and triplet attention in odd-indexed blocks and global attention in even-indexed blocks, together with state cross-attention for image conditioning (\cref{sec:attention}). The frozen VAE then decodes each denoised slot into a part mesh, and joint parameters are derived analytically from the rigid transformation between each paired rest/open part mesh (\cref{sec:urdf}).

\subsection{Analysis: Articulated Parts as Emergent Slots}
\label{sec:slot}

We analyze two failure modes of naive multi-part generation that motivate our structured slot design.

\myheading{Failure Mode I: Independent per-state generation.} Running a part-aware generator (e.g., PartCrafter~\cite{partcrafter}) independently on each image yields no cross-state slot identity: the $i$-th rest slot is unrelated to the $i$-th open slot, leading to inconsistent part binding and geometry across states (row~1, \cref{fig:slot_trajectory}). This motivates joint generation with a $1+2N$ slot layout that pairs rest/open slots by index from the outset.

\myheading{Failure Mode II: Unstructured joint generation.} Jointly generating $2N$ unstructured slots without state-specific routing or disjointness constraints provides no mechanism to pair rest/open slots or distinguish the static base from moving parts. Consequently, slot clouds remain diffuse and overlapping at $t=0$, without clear part separation in either state (row~2, \cref{fig:slot_trajectory}). This motivates explicit slot roles, state-specific conditioning, cross-state communication, and geometric separation.

Both failures stem from the absence of structural pressure for slots to bind to kinematic parts. Slot Attention~\cite{locatello2020slotattention} shows that slots jointly explaining a shared input can specialize through competition without explicit supervision. We extend this principle to articulated objects with the following hypothesis:

\begin{hypothesis}[Slot-binding] \label{hyp:slotbinding} Slots constrained to jointly explain two images of the same object in different articulation states bind to the object's kinematic parts. \end{hypothesis}

Motivated by this hypothesis, ArtCrafter adopts a \textbf{$1+2N$ slot layout}: one \textbf{base slot} represents articulation-invariant content, while $N$ \textbf{rest-state slots} and $N$ \textbf{open-state slots} are paired by index to represent the same physical parts across states. Triplet attention reinforces cross-state identity within each pair, while geometric regularization promotes inter-part disjointness (\cref{sec:loss}). Together, these constraints yield consistent part binding across articulation states by $t=0$ (row~3, \cref{fig:slot_trajectory}).

Denoising starts from $1+2N$ latent slots $\mathcal{Z} = \{\vz_{\mathrm{base}},\; \vz_{\mathrm{rest}}^1, \vz_{\mathrm{open}}^1,\; \ldots,\; \vz_{\mathrm{rest}}^N, \vz_{\mathrm{open}}^N\}$ sampled from $\mathcal{N}(0,\mathbf{I})$. Following PartCrafter~\cite{partcrafter}, each slot receives a learnable \textbf{part ID embedding} $\mathbf{e}_{\mathrm{part}}^i$ and a \textbf{state embedding} $\mathbf{e}_{\mathrm{state}} \in \{\mathbf{e}_{\mathrm{rest}}, \mathbf{e}_{\mathrm{open}}\}$ encoding part identity and articulation state. The fine-tuned TripoSG DiT denoises these slots over $T=50$ steps; the following section describes how each DiT block communicates across slots and states.

\subsection{DiT Block Attention Design}
\label{sec:attention}

The standard DiT attention provides no mechanism to enforce that paired rest/open slots describe the same physical part---without explicit coupling, the two slots may evolve independently and produce meshes unrelated by any rigid transform, making the analytic joint derivation in \cref{sec:urdf} unreliable. We therefore augment the DiT backbone with structured attention that realizes the slot-binding principle.

The DiT backbone applies \textbf{local attention}, \textbf{triplet attention}, and \textbf{global attention} in an alternating fashion, together with \textbf{state cross-attention} for image conditioning:

\begin{enumerate}[noitemsep, topsep=2pt, leftmargin=*]
    \item \textbf{Local attention} refines each slot independently before inter-slot interaction:
    \begin{align}
        \tilde{\vz}^k = \mathrm{Self\text{-}Attn}(\vz^k)
        \quad \forall\, \vz^k \in \mathcal{Z}.
        \label{eq:local_attention}
    \end{align}

    \item \textbf{Triplet attention} enforces cross-state part identity by concatenating each paired rest/open slot with the shared base slot and applying self-attention over the triplet, independently for each part $i$:
    \begin{align}
        [\tilde{\vz}_{\mathrm{rest}}^i;\,
         \tilde{\vz}_{\mathrm{open}}^i;\,
         \tilde{\vz}_{\mathrm{base}}]
        =
        \mathrm{Self\text{-}Attn}
        ([\vz_{\mathrm{rest}}^i;\,
          \vz_{\mathrm{open}}^i;\,
          \vz_{\mathrm{base}}])
        \quad \forall\, i \in \{1,\ldots,N\}.
        \label{eq:triplet_attention}
    \end{align}
    Each part slot is thereby coupled to its paired counterpart and anchored through the shared base slot, enforcing the cross-state geometric consistency required for analytic joint derivation.

    \item \textbf{State cross-attention} grounds each slot in its respective image---rest slots to $\mathbf{f}_{\mathrm{rest}}$, open slots to $\mathbf{f}_{\mathrm{open}}$, and the base slot to both---preventing cross-state appearance leakage:
    \begin{align}
        \tilde{\vz}^k
        &= \mathrm{Cross\text{-}Attn}(\vz^k, \mathbf{f}_{\mathrm{rest}})
        \quad \forall\, \vz^k \in
        \{\vz_{\mathrm{base}}\} \cup
        \{\vz_{\mathrm{rest}}^i\}_{i=1}^N, \\
        \tilde{\vz}^k
        &= \mathrm{Cross\text{-}Attn}(\vz^k, \mathbf{f}_{\mathrm{open}})
        \quad \forall\, \vz^k \in
        \{\vz_{\mathrm{base}}\} \cup
        \{\vz_{\mathrm{open}}^i\}_{i=1}^N.
        \label{eq:cross_attention}
    \end{align}

    \item \textbf{Global attention} integrates all $1+2N$ slots into a globally coherent representation:
    \begin{align}
        \tilde{\mathcal{Z}}
        =
        \mathrm{Self\text{-}Attn}(\mathcal{Z}).
        \label{eq:global_attention}
    \end{align}
\end{enumerate}

Together, these attention mechanisms provide per-slot refinement, cross-state part coupling, state-specific image conditioning, and global inter-part coherence throughout denoising.

\subsection{Geometric Regularization}
\label{sec:loss}

Global attention allows latent tokens across parts to attend to each other for coherence, but does not prevent spatial overlap between generated parts -- the FM-II failure mode in \cref{fig:slot_trajectory}. We introduce a repulsion loss that penalizes inter-part overlap directly in 3D point space, applied during training alongside the flow matching objective.

\textbf{Point decoder $\phi$.} To measure inter-part overlap during training, we train a lightweight point decoder $\phi$ to reconstruct a part point cloud from its latent tokens. Initialized from the TripoSG VAE decoder, $\phi$ replaces the final cross-attention layer with a lightweight MLP that predicts 3D coordinates from the self-attended latent set. We train $\phi$ with Chamfer loss on the same data as the DiT and then freeze it. During DiT training, $\phi$ is applied only at low-noise timesteps, where $\hat{\vz}_{0|t}$ provides a reliable estimate of the clean latent.

\textbf{Repulsion loss.}
At each denoising step, we decode the predicted clean latent $\hat{\vz}_{0|t} = \vz_t - (1-t)\, \vv_\theta(\vz_t, t)$ through $\phi$ to obtain per-slot point clouds $P_i = \phi(\hat{\vz}_{0|t}^{i})$. For each pair of distinct parts $(i,j)$ at the same state, we penalize spatial overlap via a margin-based Chamfer Distance loss:
\begin{gather}
    d(P_i, P_j) = \frac{1}{|P_i|} \sum_{p \in P_i} \max\!\left(0,\, \gamma - \min_{q \in P_j} \|p - q\|\right)^2 + \frac{1}{|P_j|} \sum_{q \in P_j} \max\!\left(0,\, \gamma - \min_{p \in P_i} \|q - p\|\right)^2, \\
    \mathcal{L}_{\mathrm{rep}} = \frac{1}{\binom{N}{2}} \sum_{(i, j) \in \binom{[N]}{2}} d(P_i, P_j),
\end{gather}
where $\gamma > 0$ is the margin threshold. Gradients flow back through the frozen $\phi$ to the part latents, coupling the abstract latent space to geometric disjointness. The progressive part separation visible in \cref{fig:slot_trajectory} illustrates its effectiveness.

\textbf{Total training objective.}
ArtCrafter is trained with rectified flow matching~\cite{lipman2022flow}, which transports a Gaussian distribution to the data distribution along a linear trajectory. Given GT part latents $\mathcal{Z}_0 = \{\vz_i\}_{i=1}^{1+2N}$ and conditioning images $\mathbf{f}_{\mathrm{open}}, \mathbf{f}_{\mathrm{rest}}$, we perturb with Gaussian noise $\boldsymbol{\epsilon} \sim \mathcal{N}(0,\mathbf{I})$ at level $t$, yielding $\mathcal{Z}_t$. The model predicts the target velocity $\boldsymbol{\epsilon} - \mathcal{Z}_0$:
\begin{equation}
  \mathcal{L}_{\mathrm{flow}} = \mathbb{E}_{\mathcal{Z},\boldsymbol{\epsilon},t}
    \left[\bigl\|(\boldsymbol{\epsilon} - \mathcal{Z}_0)
           - \mathbf{v}_\theta(\mathcal{Z}_t, t, \mathbf{f}_{\mathrm{open}}, \mathbf{f}_{\mathrm{rest}})\bigr\|^2\right], \quad \mathcal{Z}_t = t\mathcal{Z}_0 + (1-t)\boldsymbol{\epsilon},
\end{equation}
where the noise level $t$ is shared across all $2N+1$ slots. The total objective is:
\begin{equation}
  \mathcal{L} = \mathcal{L}_{\mathrm{flow}} + \lambda_{\mathrm{rep}}\,\mathcal{L}_{\mathrm{rep}},
  \label{eq:total_loss}
\end{equation}
with $\lambda_{\mathrm{rep}} = 0.5$. The point decoder $\phi$ remains frozen throughout training; the regularization biases the denoising trajectory without modifying the pretrained backbone.

%% ── §3.5 Analytic URDF Derivation ────────────────────────────────────────────
\subsection{Analytic Joint Derivation}
\label{sec:urdf}

For each geometrically consistent rest/open mesh pair $(\mathcal{M}_{\mathrm{rest}}^i, \mathcal{M}_{\mathrm{open}}^i)$, we recover the rigid transform $\mathbf{T}_i$ through global and local registration. For global registration, we extract mesh vertices as point clouds, downsample via voxelization at multiple resolutions, and compute Fast Point Feature Histograms (FPFH)~\cite{rusu2009fast} for feature matching. RANSAC over the feature correspondences yields a coarse transformation $\mathbf{T}_i' \in \mathbb{R}^{4 \times 4}$. For local registration, starting from $\mathbf{T}_i'$, Iterative Closest Point (ICP) refines the alignment to produce the final rigid transformation $\mathbf{T}_i$. Finally, from $\mathbf{T}_i = [\mathbf{R}_i \mid \mathbf{t}_i]$, we determine joint type by thresholding the rotation angle extracted from $\mathbf{R}_i \in \mathbb{R}^{3\times 3}$: large rotation angle indicates a revolute joint, otherwise it is a prismatic joint. Joint axis direction, origin, and motion range follow directly from the decomposition. The resulting part meshes and joint parameters are assembled into a physics-executable URDF file.
\section{Experiments}
\label{sec:experiments}

\subsection{Experimental Setup}

\myheading{Implementation details.} 
We train and evaluate on the dataset from URDF-Anything+~\cite{urdfanythingplus}, sourced from PartNet~\cite{partnetdataset}, PartNet-Mobility~\cite{partnetmobilitydataset}, and PhysX3D~\cite{physx3ddataset} with data cleaning and filtering, comprising 2{,}314 training, 120 validation, and 180 test objects. For each object, we render a rest-state and open-state image from a fixed frontal viewpoint alongside their ground-truth part meshes. To evaluate generalization beyond the URDF-Anything+ dataset, we additionally curate an OOD test set sourced from SINGAPO~\cite{liu2024singapo} and Artiverse~\cite{iliash2026artiverse} that comprises more complex, out-of-distribution objects. Dataset statistics, preprocessing details, and the OOD evaluation protocol are provided in the supplementary.

In ArtCrafter, we fine-tune the TripoSG~\cite{triposg} DiT backbone with the VAE frozen. Each of the $1+2N$ slots has $L=1{,}024$ tokens and latent dimension $d=64$. Training uses AdamW with cosine schedule, peak learning rate $1\times10^{-5}$, global batch size 8, for 14{,}450 iterations on 4 H100 GPUs. The repulsion margin is $\gamma=0.05$ and regularization weight $\lambda_{\mathrm{rep}}=0.5$. For VLM-prediction of number of parts $N$, we employ API from OpenAI GPT models~\cite{singh2025openai}. Further compute requirements can be found in the supplementary.

\myheading{Metrics.}
We evaluate geometry and articulation separately. For geometry, we report \textbf{Chamfer Distance} (CD) and \textbf{F-Score} (F1, threshold $0.02$) at both object and part level, where part-level matching uses Hungarian assignment on bounding box centroids. For articulation, we report \textbf{Joint Axis Error} (angular difference in degrees), \textbf{Joint Origin Error} (shortest Euclidean distance in meters), and \textbf{Joint Limit Error} (absolute motion range difference). 
We also employ another penalized version of the standard geometry and joint metrics above, where redundant generation of or missing part geometries and articulation would be penalized in order to truely assess the geometric and kinematic quality of generated articulated objects. 
All geometry metrics are evaluated at the rest state. More details about metrics and evaluation protocol can be found in the supplementary.

\myheading{Baselines.}
We evaluate two versions of ArtCrafter (with the input number of movable parts $N$ either ground-truth or generated by a VLM from the two input images) directly against DIPO~\cite{dipo}, which shares the same dual-state image setting. We additionally report representative retrieval-based (URDFormer~\cite{urdfformer}), optimization-based (FreeArt3D~\cite{freeart3d}), and generative (PhysX-Anything~\cite{physxanything}) methods to situate ArtCrafter within the broader landscape of articulated asset creation. For these methods, we align the inputs with our evaluation setting while accommodating each method’s input requirements: FreeArt3D receives the same rest/open image pair as ArtCrafter, whereas URDFormer and PhysX-Anything receive a single image.

\begin{table}[t]
  \caption{
    Quantitative results on the URDF-Anything+~\cite{urdfanythingplus} test set. Gray rows denote reference methods under different input settings. Our ArtCrafter with input number of movable parts $N$ being either ground-truth or generated by VLM. $^\dagger$~Results are reported in~\cite{urdfanythingplus}. Bold indicates (co-)best, and underline indicates second best.
  }
  \label{tab:main}
  \centering
  \footnotesize
  \setlength{\tabcolsep}{4pt}

  \begin{tabular}{@{}l cccc cccc@{}}
    \toprule

    \multirow{3}{*}{\textbf{Method}}
      & \multicolumn{4}{c}{\textbf{Geometry}}
      & \multicolumn{4}{c}{\textbf{Kinematics}} \\

    \cmidrule(lr){2-5}
    \cmidrule(lr){6-9}

      & \multicolumn{2}{c}{\textbf{Whole-Object}}
      & \multicolumn{2}{c}{\textbf{Part}}
      & \multirow{2}{*}{\textbf{Axis Err.}$\downarrow$}
      & \multirow{2}{*}{\textbf{Origin Err.}$\downarrow$}
      & \multirow{2}{*}{\textbf{Limit Err.}$\downarrow$} \\

    \cmidrule(lr){2-3}
    \cmidrule(lr){4-5}

      & \textbf{CD}$\downarrow$
      & \textbf{F1}$\uparrow$
      & \textbf{CD}$\downarrow$
      & \textbf{F1}$\uparrow$
      & & & & \\

    \midrule

    \rowcolor[gray]{0.85}
    URDFormer$^\dagger$~\cite{urdfformer}
      & 0.286
      & 0.413
      & 0.354
      & 0.389
      & 1.071
      & 0.504
      & 0.882 \\

    \rowcolor[gray]{0.85}
    FreeArt3D~\cite{freeart3d}
      & 0.113
      & 0.426
      & 0.078
      & 0.413
      & 0.668
      & 0.330
      & 1.662 \\

    \rowcolor[gray]{0.85}
    PhysX-Anything$^\dagger$~\cite{physxanything}
      & 0.068
      & 0.715
      & 0.076
      & 0.697
      & 0.191
      & 0.103
      & 0.532 \\

    \midrule

    DIPO~\cite{dipo}
      & \underline{0.047}
      & 0.289
      & \underline{0.044}
      & 0.276
      & 0.309
      & 0.107
      & 1.158 \\

    \textbf{ArtCrafter (w/ VLM)}
      & \textbf{0.004}
      & \underline{0.720}
      & \textbf{0.007}
      & \underline{0.698}
      & \underline{0.283}
      & \textbf{0.087}
      & \underline{0.653} \\

    \textbf{ArtCrafter (w/ GT)}
      & \textbf{0.004}
      & \textbf{0.721}
      & \textbf{0.007}
      & \textbf{0.700}
      & \textbf{0.279}
      & \underline{0.089}
      & \textbf{0.621} \\

    \bottomrule
  \end{tabular}

  %\vspace{-10pt}
\end{table}

\begin{table*}[t]
  \centering
  \caption{%
    Quantitative results on the OOD test set.
    Gray rows denote reference methods under different input settings.
    Bold indicates (co-)best, and underline indicates second best.
  }
  \label{tab:ood_results_simple}
  \footnotesize
  \setlength{\tabcolsep}{4pt}

  \begin{tabular}{@{}l cccc cccc@{}}
    \toprule

    \multirow{3}{*}{\textbf{Method}}
      & \multicolumn{4}{c}{\textbf{Geometry}}
      & \multicolumn{4}{c}{\textbf{Kinematics}} \\

    \cmidrule(lr){2-5}
    \cmidrule(lr){6-9}

      & \multicolumn{2}{c}{\textbf{Whole-Object}}
      & \multicolumn{2}{c}{\textbf{Part}}
      & \multirow{2}{*}{\textbf{Axis Err.}$\downarrow$}
      & \multirow{2}{*}{\textbf{Origin Err.}$\downarrow$}
      & \multirow{2}{*}{\textbf{Limit Err.}$\downarrow$} \\

    \cmidrule(lr){2-3}
    \cmidrule(lr){4-5}

      & \textbf{CD}$\downarrow$
      & \textbf{F1}$\uparrow$
      & \textbf{CD}$\downarrow$
      & \textbf{F1}$\uparrow$
      & & & & \\

    \midrule

    \rowcolor[gray]{0.85}
    PhysX-Anything~\cite{physxanything}
      & 0.076
      & 0.274
      & 0.027
      & 0.382
      & 1.148
      & 0.542
      & 1.312 \\

    \rowcolor[gray]{0.85}
    FreeArt3D~\cite{freeart3d}
      & 0.131
      & 0.135
      & 0.101
      & 0.163
      & 0.831
      & 0.419
      & 1.182 \\

    \midrule

    DIPO~\cite{dipo}
      & 0.071
      & 0.158
      & 0.072
      & \underline{0.179}
      & 1.059
      & 0.480
      & \textbf{0.906} \\

    \textbf{ArtCrafter (w/ VLM)}
      & \underline{0.047}
      & \underline{0.352}
      & \underline{0.052}
      & \textbf{0.340}
      & \underline{0.918}
      & \underline{0.463}
      & \underline{1.308} \\

    \textbf{ArtCrafter (w/ GT)}
      & \textbf{0.045}
      & \textbf{0.353}
      & \textbf{0.051}
      & \textbf{0.340}
      & \textbf{0.914}
      & \textbf{0.430}
      & 1.343 \\

    \bottomrule
  \end{tabular}

  \vspace{-10pt}
\end{table*}

\begin{figure}[t]
    \centering    \includegraphics[width=.95\textwidth]{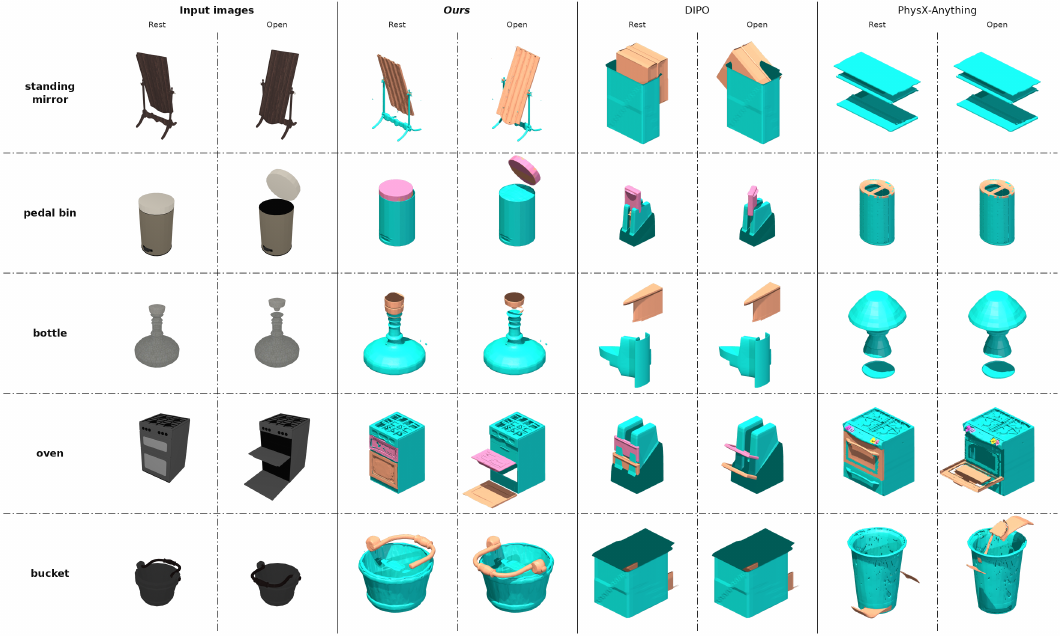}
    \vspace{-5pt}
    \caption{
    Qualitative results on URDF-Anything+~\cite{urdfanythingplus} and OOD test sets. Each row shows rest and open-state renders for the input image(s) on the left, and colors denote part identity.
    ArtCrafter produces clean part boundaries and geometry quality aligned with the inputs.
    DIPO predicts plausible kinematic behavior for certain examples (two doors of oven) but its retrieved meshes are limited on unseen objects.
    PhysX-Anything (single image) can produce aligned geometries but struggle with fine-grained details (e.g. pedal bin, oven) and could hallucinate on rare objects (mirror).
    }
    \label{fig:main_qualitative}
    \vspace{-10pt}
\end{figure}

% ── §4.2 Comparison with Prior Approaches ─────────────────────────────────────
\subsection{Main Results}
\label{sec:comparison}

\textbf{Quantitative results.} \cref{tab:main} reports geometry and articulation results on the URDF-Anything+ test set for ArtCrafter using either VLM-estimated or ground-truth (GT) number of movable parts $N$. The VLM variant remains competitive with the GT setting, demonstrating robustness to automatically estimated $N$. Under the same dual-state image setting, both variants outperform DIPO on geometry and joint metrics. We additionally report URDFormer, FreeArt3D, and PhysX-Anything as reference methods under different input settings. Inference time of these methods are reported in \cref{tab:inference_time} in the supplementary.

\textit{Geometry.} Compared directly with DIPO, the two variants improve F1 by over $40\%$ while achieving remarkably low whole-object/part CD values both with VLM-estimated $N$ and with GT $N$. Despite retrieving meshes from a fixed database, DIPO struggles to match the input geometry, whereas ArtCrafter directly synthesizes part meshes. For reference, ArtCrafter also achieves slightly better geometry quality than PhysX-Anything while requiring substantially less inference time (\cref{tab:inference_time}).

\textit{Articulation.} The VLM variant achieves nearly identical joint precision as GT, with some degradation in joint axis and limits estimation. Compared directly with DIPO, ArtCrafter is able to obtain better performance across both three joint metrics. DIPO joint prediction is unmatched by the retrieved geometry while ArtCrafter ensures both geometry and joint quality. Results under different settings provide additional context, with both ArtCrafter variants achieving articulation quality comparable to the single-image PhysX-Anything in terms of joint origin error and limits error.

\cref{tab:ood_results_simple} shows that on unseen objects, ArtCrafter obtains reasonably low CDs while achieving F-scores around $0.35$ for geometry quality that are significantly higher than those of DIPO, and shows competitive articulation prediction relative to other methods on most joint metrics.

\textbf{Qualitative results.}
\cref{fig:main_qualitative} shows representative outputs across methods. ArtCrafter produces geometrically accurate, part-separated meshes with clean boundaries across diverse categories. PhysX-Anything, conditioned on a single image without direct motion evidence, exhibits incorrect part decomposition in several examples while presenting poor geometry alignment for the standing mirror and bucket objects. Against DIPO, ArtCrafter produces substantially better quality on challenging categories where retrieval seems to fall short due to unrecognized part geometries during testing; on a common instance such as oven (row~4), DIPO predicts correct articulation behavior but cannot retrieve aligned part meshes while ArtCrafter produces a geometrically and kinematically correct result. We further show qualitative demonstration of ArtCrafter and other methods on more challenging out-of-distribution and in-the-wild inputs in the supplementary.

\begin{table}[t]
  \centering
  \caption{%
    Ablation for ArtCrafter's stage 1: Triplet attention and Regularized repulsion loss
      }
  \label{tab:stage1_ablation}
  \footnotesize
  \setlength{\tabcolsep}{7pt}
  \begin{tabular*}{0.95\textwidth}{@{\extracolsep{\fill}}l cc c cccc@{}}
    \toprule
    & \multicolumn{2}{c}{\textbf{Cross-state part enforcement}}
    & \multirow{2}{*}{\textbf{Geo.\ Reg.}}
    & \multicolumn{2}{c}{\textbf{Whole-Object}}
    & \multicolumn{2}{c}{\textbf{Part}} \\
    \cmidrule(lr){2-3}\cmidrule(lr){5-6}\cmidrule(lr){7-8}
    \textbf{Variant}
      & \textbf{Triplet attn} & \textbf{Pair-wise Attn} &
      & CD$\downarrow$ & F1$\uparrow$
      & CD$\downarrow$ & F1$\uparrow$ \\
    \midrule
    Base
      & & &
      & 0.022 & 0.579 & 0.013 & 0.608 \\
    + Geo.\ Reg.
      & & & $\checkmark$
      & \textbf{0.014} & 0.583 & 0.012 & 0.614 \\
    + Triplet attn
      & $\checkmark$ & &
      & 0.015 & 0.609 & 0.011 & 0.639 \\
    + Pair-wise Attn
      & & $\checkmark$ &
      & 0.017 & 0.580 & 0.013 & 0.592 \\
    \midrule
    \textbf{Ours}
      & $\checkmark$ & & $\checkmark$
      & \textbf{0.014} & \textbf{0.626} & \textbf{0.010} & \textbf{0.646} \\
    \bottomrule
  \end{tabular*}
  \vspace{-10pt}
\end{table}

\begin{table}[t]
  \centering
  \caption{Ablation for ArtCrafter's stage 2: Joint prediction by MLP head and analytic registration}
  \label{tab:stage2_ablation}
  \footnotesize
  \setlength{\tabcolsep}{7pt}
  \begin{tabular}{@{}lcccc@{}}
    \toprule
    \textbf{Method}
      & \textbf{Axis Err.}$\downarrow$
      & \textbf{Origin Err.}$\downarrow$
      & \textbf{Limit Err.}$\downarrow$ \\
    \midrule

    ArtCrafter (Learned Joint Head)
      & 1.049
      & 0.322
      & 2.142 \\

    ArtCrafter (Registration)
      & \textbf{0.181}
      & \textbf{0.082}
      & \textbf{0.593} \\

    \bottomrule
  \end{tabular}
  \vspace{-10pt}
\end{table}

\begin{figure}[t]
    \centering    \includegraphics[width=.95\textwidth]{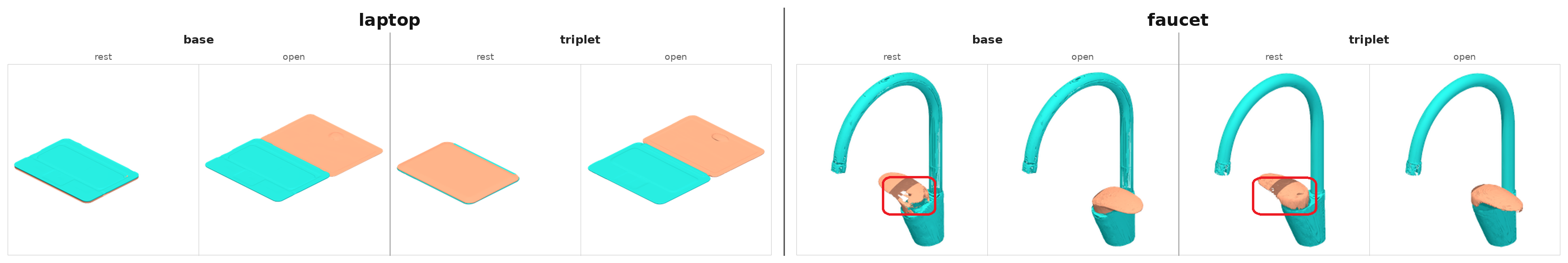}
    \vspace{-5pt}
    \caption{%
    Ablation of triplet attention, where colors denote part identity. Comparing with the base version, the triplet version enables the base and rest/ open parts to be more coherent in terms of spatial position (the laptop base and lid) and geometry quality (the faucet knob between rest and open). 
    }
    \label{fig:triplet_ablate}
    \vspace{-10pt}
\end{figure}

% ── §4.4 Applications ─────────────────────────────────────────────────────────

\subsection{Ablation Study}
\label{sec:ablation}

\cref{tab:stage1_ablation} and \cref{tab:stage2_ablation} report ablation results for both stages of ArtCrafter on the URDF-Anything+ validation set. In \cref{tab:stage1_ablation}, each row adds one component to the previous setting; since articulation parameters are derived analytically from the generated meshes, geometry metrics directly reflect joint quality and serve as the primary ablation factors. In \cref{tab:stage2_ablation}, we replace the original analytic registration approach with a learned network on the denoised open latent tokens from DiT backbone. We provide more qualitative demonstrations in the supplementary.

\textbf{Necessity of triplet attention}
By introducing triplet attention to the base architecture, we manage to improve the F1 score of approximately $3\%$ at both whole-object and part-level over the base variant. This confirms our hypothesis that without triplet attention, paired slots evolve independently and unrelated meshes that are not well connected by a rigid transformation. Therefore, cross-state coupling is a prerequisite for geometric consistency across articulation states. Two ablated examples are provided in ~\cref{fig:triplet_ablate}.

\textbf{Effectiveness of base anchoring over pairwise attention.} Without the base slot, each rest/open pair attends only to itself, allowing pairs to become internally consistent while remaining misaligned with other parts. The shared base provides an object-level reference across all pairs, enforcing global spatial coherence. This explains why pairwise attention improves whole-object geometry but not part-level quality, while full triplet attention improves both.

\textbf{Importance of geometric regularization for part disjointness.} Without geometric regularization, decoded part meshes may overlap and degrade geometry quality. The repulsion loss $\mathcal{L}_{\mathrm{rep}}$ improves both CD and F1 by enforcing spatial disjointness between parts. Combined with triplet attention, the final model improves F1 by approximately $4\%$ over the base variant while reducing CD, demonstrating the complementary roles of cross-state consistency and inter-part disjointness.

\textbf{Effectiveness of analytic registration for joint derivation. } In \cref{tab:stage2_ablation}, analytic joint derivation can predict joint axis, origin, and limits more precisely and in a more interpretable way compared to learned joint heads, which would struggle under limited articulation data in practice.

% \vspace{-10pt}
\section{Conclusion}
\label{sec:conclusion}

We presented ArtCrafter, a two-stage feed-forward framework for generating articulated 3D assets from a pair of rest/open images. In the first stage, a $1+2N$ slot DiT jointly generates part-level meshes across both articulation states, with triplet attention enforcing cross-state geometric consistency and $\mathcal{L}_{\mathrm{rep}}$ promoting inter-part disjointness. In the second stage, joint parameters are derived analytically from the rigid transformations between paired rest/open meshes. Experiments on URDF-Anything+ demonstrate substantial improvements over DIPO under the same dual-state setting, while ArtCrafter remains competitive with reference methods. Evaluation on the more complex OOD benchmark further demonstrates generalization to unseen object categories and structures.

\begin{ack}
% TODO: Add acknowledgments and funding disclosure here.
% Note: omit this section in the anonymized submission.

This work was funded by Qualcomm Technologies, Inc. The authors declare no competing interests.

\end{ack}

\clearpage
\bibliographystyle{abbrv}
\bibliography{references}

%%%%%%%%%%%%%%%%%%%%%%%%%%%%%%%%%%%%%%%%%%%%%%%%%%%%%%%%%%%%

% \appendix

\clearpage
\appendix

\section{Limitations and Broader Impact}
\label{sec:limitations}

\textbf{Limitations.} 
ArtCrafter currently generates geometry without part textures. The two-image input assumes availability of a clearly open-state observation, which holds for most household objects but may not generalize to novel articulation types. Analytic URDF derivation assumes rigid parts, excluding flexible or deformable components. Training and evaluation are conducted on the URDF-Anything+ dataset; generalization to open-vocabulary articulated objects with complex multi-joint kinematics remains to be studied.

\textbf{Broader impact.}
ArtCrafter lowers the barrier to sim-ready articulated asset creation, with potential to accelerate robot learning research at scale. Generated assets inherit the coverage biases of URDF-Anything+; users should validate generated URDFs in simulation before deployment in safety-critical settings.

\section{Additional Results}
\label{sec:appendix}
% TODO: Add supplementary material, proofs, and additional experiments here.

\subsection{Detailed specification of datasets and preprocessing}

\subsubsection{URDF-Anything+ dataset}

\textbf{Detailed statistics. } We provide further details regarding total number of parts and number of joints per type in \cref{tab:uap_stats}. Among the three sources, $418$ objects come from the PartNet-Mobility dataset, while the remaining $2196$ objects are sourced from PartNet and PhysX-3D datasets with more complex categories such as Faucet, Scissors, Knife, and more diverse objects per category of PartNet-Mobility. (Since PhysX-3D is sourced from PartNet but with more detailed physical information, we group those objects to both of these sources.) In \cref{tab:uap_num_parts} of this rebuttal, we further provide statistics about the number of objects that have a specific number of parts in the URDF-Anything+ dataset, indicating that this dataset contains objects with different number of parts, and the most complex object possesses 5 parts.

\begin{table}[t]
  \centering
  \caption{Detailed statistics of the URDF-Anything+ dataset.}
  \label{tab:uap_stats}
  \footnotesize
  \setlength{\tabcolsep}{8pt}
  \begin{tabular}{@{}lc@{}}
    \toprule
    \textbf{Statistic} & \textbf{Count} \\
    \midrule
    Objects          & 2{,}614 \\
    Categories       & 10 \\
    Parts            & 5{,}990 \\
    Joints           & 3{,}376 \\
    Revolute joints  & 3{,}217 \\
    Prismatic joints & 159 \\
    \bottomrule
  \end{tabular}
\end{table}
\begin{table}[t]
  \centering
  \caption{Distribution of number of objects per part-count in URDF-Anything+ dataset.}
  \label{tab:uap_num_parts}
  \footnotesize
  \setlength{\tabcolsep}{8pt}
  \begin{tabular}{@{}lcc@{}}
    \toprule
    \textbf{Number of parts} & \textbf{Count} & \textbf{Proportion} \\
    \midrule
    2 & 1{,}962 & 75.1\% \\
    3 & 542     & 20.7\% \\
    4 & 109     & 4.2\% \\
    5 & 1       & 0.04\% \\
    \midrule
    \textbf{TOTAL} & \textbf{2{,}614} & \textbf{100\%} \\
    \bottomrule
  \end{tabular}
\end{table}

\textbf{Preprocessing.} The URDF-Anything+~\cite{urdfanythingplus} dataset lacks the paired rest/open images and meshes required by our setting. We therefore apply forward kinematics to obtain each object's rest and open meshes, center and normalize them to the $[-1,1]^3$ cube, and render the corresponding RGB input images from a fixed viewpoint. The resulting meshes serve as ground truth for geometry evaluation, while joint parameters are extracted from the URDFs and transformed into a unified world coordinate frame for articulation evaluation. We split the preprocessed dataset into train/validation/test sets with per-category counts reported in \cref{tab:uap_split}.

\begin{table}[t]
\centering
\begin{tabular}{lcccc}
\toprule
\textbf{Category} & \textbf{Train} & \textbf{Val} & \textbf{Test} & \textbf{Total} \\
\midrule
Dishwasher           & 200 & 10 & 14 & 224 \\
Display              & 253 & 13 & 24 & 290 \\
Door                 & 40  & 2  & 5  & 47  \\
Faucet               & 575 & 30 & 49 & 654 \\
Knife                & 23  & 1  & 3  & 27  \\
Laptop               & 408 & 21 & 41 & 470 \\
Microwave            & 234 & 12 & 22 & 268 \\
Refrigerator         & 205 & 11 & 21 & 237 \\
Scissors             & 28  & 2  & 1  & 31  \\
StorageFurniture     & 348 & 18 & 0  & 366 \\
\midrule
\textbf{TOTAL}             & 2314 & 120 & 180 & 2614 \\
\bottomrule
\end{tabular}
\vspace{5pt}
\caption{Train-val-test split statistics for the preprocessed URDF-Anything+ dataset.}
\label{tab:uap_split}
\end{table}

\subsubsection{OOD test set}

We select articulated objects from the released SINGAPO~\cite{liu2024singapo} and Artiverse dataset~\cite{iliash2026artiverse} whose categories are unseen and semantically distinct from the URDF-Anything+ categories in \cref{tab:uap_split}. This ensures that the OOD objects have fundamentally different geometries, functionalities, and articulation structures from those in URDF-Anything+. New categories include Bucket, Eyeglasses, Lamp, Pedal Bin, Mirror, Switch, and Clock. \cref{tab:artiverse_stats} and \cref{tab:artiverse_num_parts} summarize this OOD test set.

\begin{table}[t]
  \centering
  \caption{Detailed statistics of the OOD test set.}
  \label{tab:artiverse_stats}
  \footnotesize
  \setlength{\tabcolsep}{8pt}
  \begin{tabular}{@{}lc@{}}
    \toprule
    \textbf{Statistic} & \textbf{OOD test set} \\
    \midrule
    Objects       & 187 \\
    Categories    & 46 \\
    Parts         & 537 \\
    Joints        & 350 \\
    Rev. joints   & 280 \\
    Pris. joints  & 70 \\
    \bottomrule
  \end{tabular}
\end{table}

\begin{table}[t]
  \centering
  \caption{Distribution of number of objects per part-count in Artiverse OOD test set.}
  \label{tab:artiverse_num_parts}
  \footnotesize
  \setlength{\tabcolsep}{8pt}
  \begin{tabular}{@{}lcc@{}}
    \toprule
    \textbf{Num parts} & \textbf{Count} & \textbf{\%} \\
    \midrule
    2 & 72 & 38.5\% \\
    3 & 77 & 41.2\% \\
    4 & 28 & 15.0\% \\
    5 & 10 & 5.3\% \\
    \midrule
    \textbf{TOTAL} & \textbf{187} & \textbf{100\%} \\
    \bottomrule
  \end{tabular}
\end{table}

\subsubsection{Mesh post-processing}

The predicted part meshes produced by the model often contain artifacts such as small disconnected components and spurious geometric fragments due to imperfect reconstruction or noise. To address this, we perform a post-processing step that removes isolated components based on their face count relative to the entire mesh. Specifically, for each predicted mesh, we identify connected components and compute their face counts, and remove those whose number of faces falls below a predefined percentage threshold of the mesh’s total face count. This effectively filters out insignificant fragments that are unlikely to correspond to meaningful object parts. To ensure robustness, we adopt an adaptive strategy: if removing small components at the initial threshold results in empty meshes, the threshold is iteratively halved until all parts retain non-zero geometry. After removal, any unreferenced vertices are discarded to maintain mesh consistency. This procedure preserves the dominant structural components while eliminating redundant small artifacts, leading to cleaner and more stable geometry for downstream processing.

\subsection{Metrics}

Regarding the evaluation protocol, we employ standard metrics for generated geometry and predicted joint parameters following Articulate-Anything~\cite{liu2024articulateanything}. For the more complex Artiverse OOD test set, where methods may under- or over-predict the number of parts and joints, we additionally report penalized variants of the applicable part-level geometry and articulation metrics to account for such structural prediction errors.

\subsubsection{Geometry metrics}

We provide the following mathematical formulas of the geometry metrics we used:

$$CD(P_{pred}, P_{gt}) = \frac{1}{|P_{pred}|} \sum_{p \in P_{pred}} \min_{g \in P_{gt}} |p-g|_2^2 + \frac{1}{|P_{gt}|} \sum_{g \in P_{gt}} \min_{p \in P_{pred}} |g-p|_2^2$$

where we sampled $204,800$ points uniformly from both the predicted and ground-truth meshes to obtain $P_{pred}$ and $P_{gt}$, respectively. For calculating the F-score, we used the following formulas to compute precision and recall in advance with predefined threshold $\tau = 0.02$ from URDF-Anything+:

$$Precision = \frac{|{p \in P_{pred} \mid d(p,P_{gt}) < \tau}|}{|P_{pred}|}, \qquad \tau = 0.02$$

$$Recall = \frac{|{g \in P_{gt} \mid d(g,P_{pred}) < \tau}|}{|P_{gt}|}, \qquad \tau = 0.02$$

$$F1 = \frac{2 \cdot Precision \cdot Recall}{Precision + Recall}$$

During geometry evaluation at both object-level and part-level, we perform translational and uniform scale alignment between the predicted and ground-truth meshes to correctly reflect the quality of the generated meshes with the metrics above.

\subsubsection{Joint metrics}

Regarding the joint metrics, \textbf{joint axis error} measures the average angular deviation between predicted and ground-truth axis directions (in radians). \textbf{Joint origin error} measures the average distance between predicted and ground-truth axes (in meters). Finally, \textbf{joint limit error} measures the average discrepancy between predicted and ground-truth motion ranges. Detailed formulas are provided in Appendix A.1.2 of Articulate-Anything~\cite{liu2024articulateanything}.

\subsubsection{Penalized metrics}

Since the Artiverse OOD test set contains objects with more complexity in terms of shape, texture, and articulation structure compared to those from URDF-Anything+ test set, the wrong prediction of number of parts could become more severe and hinder the goal of generating articulated objects that are geometrically and kinematically aligned with inputs. In light of this problem, we employ another penalized version of the standard geometry and joint metrics we used thus far, where both the redundant generation of part geometries and articulation would be penalized in order to truely assess the geometric and kinematic quality of generated articulated objects. In particular, after Hungarian matching in part-level evaluation, we employ a penalty for the lack of or redundant predicted parts. Specifically, following established penalized metrics proposed in Particulate~\cite{particulate}, we compute each metric $D_d$ over object's predicted parts $\lbrace p_i^{\mathrm{pred}} \rbrace_{i=1}^{N}$ and GT parts $\lbrace p_j^{\mathrm{gt}} \rbrace_{j=1}^{M}$:

$$D_d = \frac{1}{2} \left( \frac{1}{N}\sum_{i=1}^{N}\tilde{d}(p_i^{pred}) + \frac{1}{M}\sum_{j=1}^{M}\tilde{d}(p_j^{gt}) \right)$$

where $\tilde{d}(p_i) = d(p_i, H(p_i))$ if (predicted or GT) part $p_i$ is matched to a (GT or predicted) part $H(p_i)$, and is otherwise set to a penalty value $\epsilon = 0$ for F-score, and half the diagonal length of the input mesh's bounding box for CD. The $d$ sets the metric and $D_d$ reports that metric with penalization for unmatched elements. For under-predicted or over-predicted joints, we apply the same formula above, treating $\lbrace a_i^{\mathrm{pred}} \rbrace_{i=1}^{N}$ and $\lbrace a_j^{\mathrm{gt}} \rbrace_{j=1}^{M}$ as the predicted and GT joints, respectively, and define:

$$\tilde{d}(a_i)=d(a_i,H(a_i))= \mathit{axis\ err},\: \text{if matched, else }\dfrac{\pi}{2}$$

In the equation above, we set the worst value for axis error for unmatched joint elements as penalties, and we exclude penalties for origin error and limits error for the sake of transparency. By this way, we can penalize both the underprediction and overprediction of part geometries and articulations, facilitating the observation of a more significant performance gap of evaluation on the Artiverse OOD test set as in \cref{tab:ood_results}.

\subsection{Joint Derivation from Rest-Open Part Meshes}

\subsubsection{Rest-Open Meshes Registration}

We estimate the joint parameters by registering the predicted rest-state part mesh to its corresponding open-state mesh using a coarse-to-fine RANSAC-ICP pipeline. 

First, both meshes are converted into point clouds. The registration is performed at multiple spatial resolutions to improve robustness. At each resolution, the point clouds are downsampled, and local FPFH geometric descriptors are extracted to support feature-based matching. A coarse global alignment is then estimated using RANSAC, which samples candidate correspondences between the two point clouds and validates rigid transformation hypotheses using geometric consistency constraints. This step provides an initial transformation that is robust to large pose differences and local minima.

Starting from this coarse alignment, point-to-point ICP is applied to the original point clouds to refine the registration and obtain the final rigid transformation $T = \begin{bmatrix} R & t \\ 0 & 1 \end{bmatrix}$, which maps the rest-state geometry to the open-state geometry. The joint parameters are then extracted from this transformation.

Since the registration quality can vary with spatial resolution, we repeat this procedure over multiple resolutions and select the final result based on the alignment between the predicted joint axis and the canonical coordinate axes. Specifically, for a normalized predicted axis $a$, we compute $\max(|a_x|,|a_y|,|a_z|)$, which measures its cosine similarity to the closest standard axis up to sign ambiguity. The candidate with the highest score is selected as the final joint estimate, encouraging stable axis predictions consistent with the normalized object coordinate system.

Regarding \textbf{registration hyperparameters}, we choose multiple voxel sizes for voxel downsampling that are fractions of part's bbox's diagonal length ($1/5$, $1/10$, $1/30$, $1/60$, $1/80$, $1/100$). We compute FPFH features with neighbor search radius of $5\times$ voxel size and search max neighbors of $100$. For RANSAC, we use maximum correspondence points-pair distance of $1.5\times$ the voxel size, the transform estimation method is point-to-point, with a maximum of $100,000$ iterations. For ICP, we use the standard point-to-point variant with maximum correspondence points-pair distance being $2\times$ the voxel size and the algorithm goes through a maximum of $2,000$ iterations.

\subsubsection{Joint derivation from Transformation}

Given a rigid transformation $T \in SE(3)$, we derive joint parameters from its rotational and translational components. The rotation matrix $R \in SO(3)$ is converted to axis-angle form using the logarithm map, yielding a rotation vector $\textbf{r} = \log(R)$ with magnitude $\theta = \|\textbf{r}\|$. If $\theta < \alpha$, the motion is classified as prismatic; otherwise, it is revolute. For a revolute joint, the axis direction is given by the normalized rotation vector $\omega = \textbf{r} / \|\textbf{r}\|$. A point on the rotation axis is obtained by solving $(I - R)p = t$ using the pseudoinverse, $p_0 = (I-R)^+t$, and the joint origin is defined by projecting the mesh centroid $c$ onto this axis, $p = p_0 + ((c-p_0).\omega)\omega$. The joint limit is represented by the rotation angle $\theta$. For a prismatic joint, the axis direction is extracted from translation as $d = t / \|t\|$, and the motion range is $\|t\|$, while the joint origin is set to the mesh centroid. This formulation follows directly from rigid body kinematics and the axis-angle representation of rotations.

\subsection{Point Decoder}

Regarding the point decoder, our goal is to train a network that can predict coordinates of 3D point cloud from VAE latents of an object’s part encoded by the VAE encoder. To utilize an available good prior for point cloud prediction from VAE latents, we reuse the VAE decoder of TripoSG~\cite{triposg} by keeping all of its self-attention layers and replacing the last cross-attention layer with a small MLP to generate 3D coordinates of the point cloud in $[-1, 1]^3$ cube. During training this additional MLP, the VAE decoder backbone is kept frozen and we only train this MLP. During training the DiT backbone of ArtCrafter, the whole network is frozen as it is used for point cloud decoding purpose only. We use the same URDF-Anything+ train set to train this point decoder so it can be used during ArtCrafter training time.

\subsection{Baselines}

We draw on the recorded results of PhysX-Anything and URDFormer in geometry quality and kinematic behavior in  URDF-Anything+~\cite{urdfanythingplus}.

\subsubsection{DIPO}

\textbf{DIPO} uses a lightweight denoising diffusion probabilistic model to generate coarse part bounding boxes and attributes including semantic part labels, joint type, joint axis, and joint limits, supported by a VLM for kinematic structure generation and a retrieval-based algorithm for fine-grained mesh retrieval from the training database. To ensure fair comparison with ArtCrafter, we trained DIPO on the URDF-Anything+ train set over the same 50 epochs as ArtCrafter, while keeping other hyperparameters of DIPO as reported in their paper and released codebase.

As the diffusion backbone of DIPO and its retrieval algorithm require semantic part labels for effective modeling and mesh retrieval while the majority of samples from the URDF-Anything+~\cite{urdfanythingplus} dataset do not include semantic labels for object parts, we specifically curate a fixed set of semantic part labels for URDF-Anything+ train set to facilitate DIPO training. We employed OpenAI GPT~\cite{singh2025openai} to infer object part labels from input prompt and the two input images. Afterwards, we obtained a unified set of $11$ semantic labels for all object parts within the URDF-Anything+ train set: \textit{base}, \textit{blade}, \textit{button}, \textit{door}, \textit{drawer}, \textit{handle}, \textit{knob}, \textit{lid}, \textit{screen}, \textit{spout}, and \textit{stand}. This fixed set of part labels is used during both training and inference, guiding the VLM component of DIPO to assign semantic labels to parts within this set and to derive the output part labels for mesh retrieval from this set in both in-distribution and out-of-distribution testing.

Regarding training hyperparameters, we employed the same mixed-precision (fp16) with training batch size of $20$, AdamW optimizer with $\beta = (0.9, 0.99)$, learning rate of $5 \times 10^{-4}$ for the image-conditioned module and $5 \times 10^{-5}$ for the base model, and other optimization parameters from the released codebase of DIPO. Training and inference are completed on NVIDIA H100 GPUs. The maximum number of parts is kept at 32 with architectural parameters remaining the same and visual features extracted from DINOv2 are cached for training and inference following implementation from DIPO. Results from DIPO are evaluated on our employed metrics as reported in the main paper.

\subsubsection{FreeArt3D}

\textbf{FreeArt3D} is a training-free pipeline: it reconstructs per-part meshes via TRELLIS, estimates joint axes via GIM feature matching with RANSAC, then refines geometry and joint parameters jointly through SDS optimization over a hash-grid neural field. In the two-state setting we use, the SDS stage runs for 2{,}000 iterations with learning rates $0.01$ for geometry and articulation, $0.1$ for pivot, and SDS weight $0.1$. No model training is performed; inference runs per object independently.

\subsubsection{ArtCrafter with learned joint heads}

\textbf{ArtCrafter with learned joint heads.} From \cref{tab:stage2_ablation}, we provide a controlled baseline to replace the original analytic registration approach with a learned network on the denoised open latent tokens. We apply categorical cross-entropy and L1 loss for joint type classification and regression, respectively. During training, such articulation supervision is only applied at low-noise timesteps and at the last $20\%$ of the training procedure to ensure the model has learned good latent representation first.

\subsection{Performance on out-of-distribution samples}

\begin{table*}[t]
  \centering
  \caption{Evaluation on the OOD test set using standard and penalized metrics. Gray rows denote reference methods under different input settings. Penalized metrics account for under- and over-prediction of object parts and joints.}
  \label{tab:ood_results}
  \scriptsize
  \setlength{\tabcolsep}{3.5pt}
  \begin{tabular}{@{}l cc cc c c cc@{}}
    \toprule
    \multirow{3}{*}{\textbf{Method}}
      & \multicolumn{4}{c}{\textbf{Part Geometry}}
      & \multicolumn{2}{c}{\textbf{Whole-Object Geometry}}
      & \multicolumn{2}{c}{\textbf{Kinematics}} \\
    \cmidrule(lr){2-5}
    \cmidrule(lr){6-7}
    \cmidrule(lr){8-9}
      & \multicolumn{2}{c}{\textbf{F-score}$\uparrow$}
      & \multicolumn{2}{c}{\textbf{CD}$\downarrow$}
      & \multirow{2}{*}{\textbf{F-score}$\uparrow$}
      & \multirow{2}{*}{\textbf{CD}$\downarrow$}
      & \multicolumn{2}{c}{\textbf{Axis Err}$\downarrow$} \\
    \cmidrule(lr){2-3}
    \cmidrule(lr){4-5}
    \cmidrule(lr){8-9}
      & \textbf{Std.} & \textbf{Pen.}
      & \textbf{Std.} & \textbf{Pen.}
      & &
      & \textbf{Std.} & \textbf{Pen.} \\
    \midrule

    \rowcolor[gray]{0.85}
    PhysX-Anything
      & 0.382 & 0.317
      & 0.027 & 0.091
      & 0.274 & 0.076
      & 1.148 & 1.341 \\

    \rowcolor[gray]{0.85}
    FreeArt3D
      & 0.163 & 0.146
      & 0.101 & 0.143
      & 0.135 & 0.131
      & 0.831 & 1.226 \\

    \midrule

    DIPO
      & \underline{0.184} & 0.157
      & 0.074 & 0.129
      & 0.158 & 0.071
      & 1.059 & 1.174 \\

    \textbf{ArtCrafter (w/ VLM)}
      & \textbf{0.340} & \underline{0.325}
      & \underline{0.052} & \underline{0.070}
      & \underline{0.352} & \underline{0.047}
      & \underline{0.918} & \underline{0.974} \\

    \textbf{ArtCrafter (w/ GT)}
      & \textbf{0.340} & \textbf{0.340}
      & \textbf{0.051} & \textbf{0.051}
      & \textbf{0.353} & \textbf{0.045}
      & \textbf{0.914} & \textbf{0.914} \\

    \bottomrule
  \end{tabular}
\end{table*}

Although the URDF-Anything+ test set contains diverse object categories, we further benchmark methods on the challenging out-of-distribution test set with more diverse categories, visually realistic textures, and complex shapes. This additional benchmark evaluates ArtCrafter's generalization to novel classes and provides broader context for other methods' capabilities on complex synthetic articulated 3D objects with visually realistic textures.

\cref{tab:ood_results} shows that ArtCrafter maintains good geometry generation and articulation prediction performance relative to other methods. In particular, on \textbf{standard} metrics, ArtCrafter obtains reasonably low CDs while achieving F-scores around $0.34$ for geometry quality, and shows good articulation prediction in terms of joint axis error relative to other methods. By \textbf{penalizing} underprediction and overprediction of parts and articulations, the performance of other methods in terms of joint axis error has decreased significantly. The large axis error increase of PhysX-Anything from $1.148$ to $1.341$ radian is due to the severe combination of underprediction and overprediction issues on the OOD test set: this test set has a total of 350 joints, but PhysX-Anything over-predicted 257 extra joints and underpredicted 129 joints simultaneously, so that has penalized the type accuracy and axis error significantly. ArtCrafter with predicted $N$ from VLM suffers less from this problem with only 47 missing joints and 9 extra joints. Part-level metrics of all methods slightly worsen compared to the non-penalized results. Note that the penalized metrics for ArtCrafter with $N$ ground-truth variant is same as that of standard metrics, for easy observation. The employed penalized metrics enable the observation of more significant performance gaps between ArtCrafter and other methods once the growing problem of wrong number of parts prediction is taken into consideration.

Broadly speaking, the results from \cref{tab:ood_results} illustrate that current methods still fall short for complex articulated objects from novel categories and diverse textures. Although ArtCrafter is trained with simple rendered input images, it still possesses reasonable generalization to unseen classes of articulated objects with visually realistic textures.

\subsection{Compute and time requirements for training and prediction}

\begin{table}[t]
  \centering
  \caption{Required time and memory during ArtCrafter's training and prediction on H100 GPU.}
  \label{tab:compute_table}
  \footnotesize
  \setlength{\tabcolsep}{8pt}
  \begin{tabular}{@{}lccc@{}}
    \toprule
      & \textbf{Time} & \textbf{RAM} & \textbf{VRAM} \\
    \midrule
    Training (4 H100s) & 3 hrs 50 mins & 20 GB & 43.2 GB \\
    Prediction (1 H100) & 22.4 sec & 10.3 GB & 10.5 GB \\
    \bottomrule
  \end{tabular}
\end{table}

\begin{table}[t]
  \centering
  \caption{Average inference time of different methods and ArtCrafter on H100 GPU.}
  \label{tab:inference_time}
  \footnotesize
  \setlength{\tabcolsep}{7pt}
  \begin{tabular}{@{}lcl@{}}
    \toprule
    \textbf{Method} & \textbf{Inference time} & \textbf{Details} \\
    \midrule
    DIPO & 12.3 sec & \\
    FreeArt3D & 606 sec & \\
    PhysX-Anything & 395 sec & \\
    \textbf{ArtCrafter (Ours)} & \textbf{22.4} sec & DINO+DiT+VAE: 18.7 sec, Registration: 3.7 sec \\
    \bottomrule
  \end{tabular}
\end{table}

In \cref{tab:compute_table}, we provide further details about the required time and memory needed during training and prediction. For \textbf{training}, we report the total required time of approximately 3 hours and 50 minutes when ArtCrafter was trained for $14,450$ iterations on 4 H100 NVIDIA GPUs with PyTorch Distributed Data Parallel (DDP) mode and per-GPU batch size of 2. The peak RAM is approximately 20 GB and the peak VRAM is approximately 43.2 GB per GPU during the entire training process. For \textbf{prediction}, ArtCrafter requires about 22.4 seconds for both geometries generation and articulation prediction, while both RAM and VRAM consumptions stay modest at 10.3 GB and 10.5 GB, respectively.

In addition, we provide the prediction time of methods in \cref{tab:inference_time}. We use the reported runtime of FreeArt3D from their paper (on the same H100 GPU) while manually measuring the average runtime of DIPO and PhysX-Anything. We measure prediction time of generating all articulated objects in the URDF-Anything+ test set and report the average inference time. As can be seen in \cref{tab:inference_time}, ArtCrafter runs approximately $27\times$ faster than FreeArt3D and $17\times$ faster than PhysX-Anything. Although DIPO is nearly twice as fast, ArtCrafter delivers substantially better geometry and articulation quality.

\subsection{More qualitative results}

\subsubsection{Performance on challenging out-of-distribution cases}

\begin{figure}[t]
    \centering    \includegraphics[width=.95\textwidth]{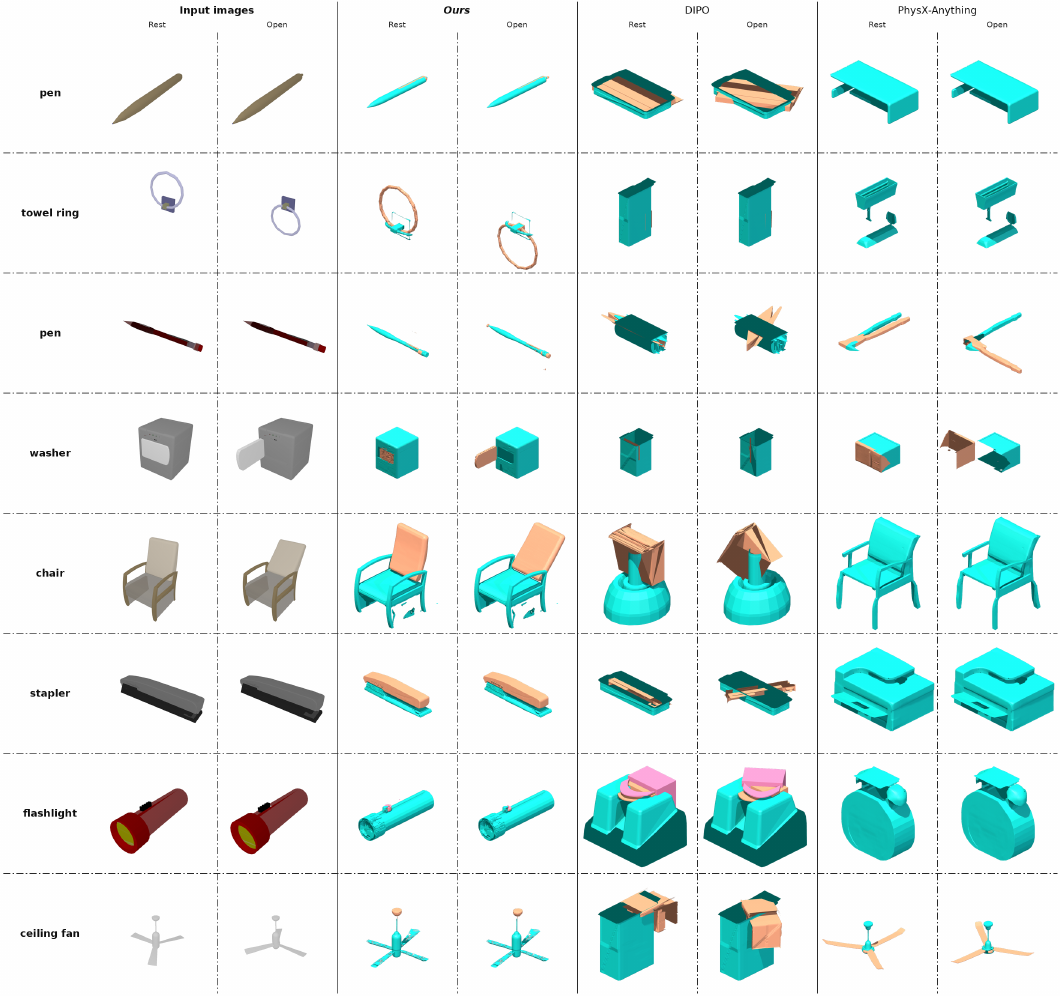}
    \vspace{-5pt}
    \caption{%
    Qualitative results of ArtCrafter and other methods on challenging out-of-distribution objects}
    \label{fig:appendix_ood_more}
    \vspace{-10pt}
\end{figure}

\cref{fig:appendix_ood_more} shows more examples of out-of-distribution articulated objects where ArtCrafter is able to generate aligned geometries with clear part boundaries and sound motion behavior, while other methods struggle with either or both geometry generation and articulation prediction.

\subsubsection{Performance on in-the-wild scenarios}

\begin{figure}[t]
    \centering    \includegraphics[width=.95\textwidth]{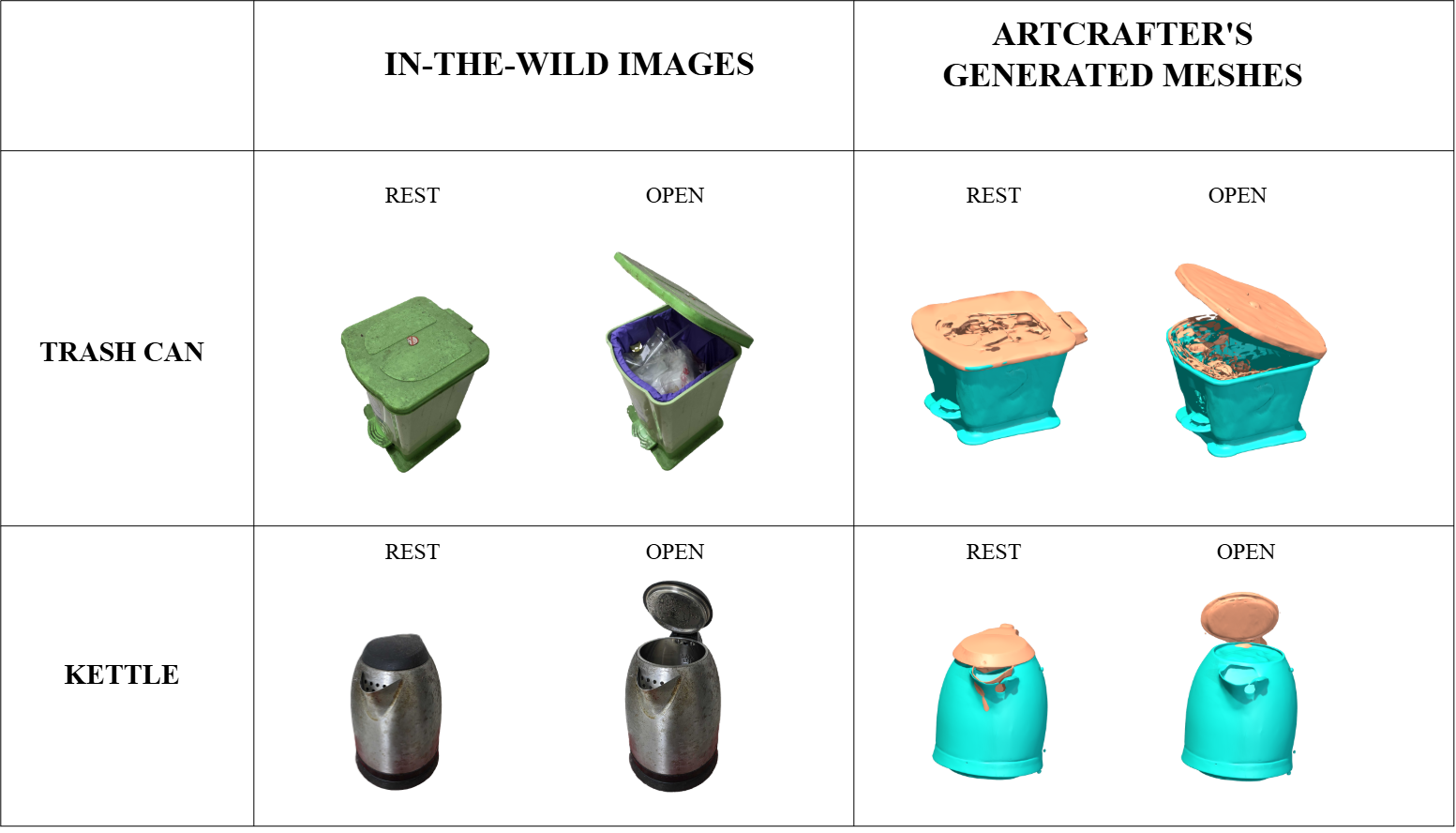}
    \vspace{-5pt}
    \caption{%
    Qualitative results of ArtCrafter on in-the-wild, unseen images}
    \label{fig:appendix_itw}
    \vspace{-10pt}
\end{figure}

To further test the robustness of ArtCrafter when encoutering completely new objects that it never saw during training, we provide the following qualitative examples where ArtCrafter is asked to generate output meshes for two unseen objects relative to URDF-Anything+ dataset: \textbf{kettle} and \textbf{trash can}. We capture real RGB images of these two objects in our own workspace and have ArtCrafter generate the complete rest and open meshes of each object. The qualitative results can be seen in \cref{fig:appendix_itw}. The qualitative results show that with completely new objects like a real kettle or a real trash can, ArtCrafter is still able to recognize and generate faithful rest and open meshes with reasonable and kinematically meaningful part decomposition. The trash can is a challenging case where the garbage inside brings some noisy information to the model and causes it to generate some redundant meshes on the rim of the body part that belong to the lid part. Nevertheless, the overall geometries of both states align well with the input images, and there is a clear separation between the body and the lid of the trash can. For the kettle, ArtCrafter can generate faithful geometries both at object-level and part-level that align well with the input images.

In terms of generalization to real-life images, to the best of our knowledge, there is currently a shortage of articulated object data captured in the wild with complete geometry and articulation ground truths that can be used for training and evaluation on our employed metrics. From the OOD test results in \cref{tab:ood_results}, there is still an existing gap for previous methods to obtain good performance on generating realistic articulated objects, and ArtCrafter is able to show certain robustness despite being trained on limited data. Thus, we believe that by training on larger and more diverse data with visually realistic textures, ArtCrafter can generalize better to in-the-wild images. We leave this as a future improvement from our current work.

\subsection{Further discussions}

\subsubsection{Further discussion on the input number of parts $N$ to ArtCrafter}

One important factor of the ArtCrafter framework is the input number of parts $N$ that determines the exact slot layout representing the object's parts at the two states. Our main results show the robustness of ArtCrafter when $N$ is given or predicted by a VLM from two-state images. We provide further information about this input $N$ below.

\textbf{Input $N$ during training and inference.} During training ArtCrafter, we employ the ground-truth $N$ as input for every object sample. Therefore, after training we use a single shared checkpoint on objects with varying numbers of movable parts $N$ and it does not require a separately trained model for each choice of $N$. During inference, $N$ only determines the number of instantiated latent slots, $1 + 2N$; all backbones' parameters are shared across different values of $N$. Therefore, the same trained model can be evaluated with any supported $N$.

\begin{table}[t]
  \centering
  \caption{Ablation results of ArtCrafter with fixed $N=5$ on the URDF-Anything+ test set.}
  \label{tab:fixed_n_ablation}
  \footnotesize
  \setlength{\tabcolsep}{6pt}
  \begin{tabular}{@{}lcccc@{}}
    \toprule
    \textbf{Method} &
    \textbf{Parts F-Score ($\uparrow$)} &
    \textbf{Parts CD ($\downarrow$)} &
    \textbf{Whole F-Score ($\uparrow$)} &
    \textbf{Whole CD ($\downarrow$)} \\
    \midrule
    ArtCrafter (with fixed $N=5$) & 0.640 & 0.00897 & 0.686 & 0.00631 \\
    ArtCrafter & 0.716 & 0.006 & 0.732 & 0.006 \\
    \bottomrule
  \end{tabular}
\end{table}

\textbf{Behavior of ArtCrafter with varying input $N$.} Regarding behavior of ArtCrafter during prediction when a different input $N$ is given than the actual number of parts of the object, ArtCrafter is still able to recognize the object's true number of parts most of the time based on visual and motion evidence from the dual-state input images. Regarding the internal slot binding mechanism, ArtCrafter will bind the slot tokens to object's parts in such a way that there will be more than one slot token binding to the same part, ensuring that all slot tokens bind to the true number of parts in the end. In order to keep a one-to-one mapping between the slot tokens and the object's number of parts, we perform an additional post-processing step of removing redundant slots binding to the same part by measuring their degree of overlap in the 3D space and thresholding, motivated by the non-max suppression technique. In the \cref{tab:fixed_n_ablation}, we provide an ablation experiment to verify this behavior by setting $N=5$, which is the max number of parts of any object in the URDF-Anything+ dataset, for all inference cases and removing the redundant slot tokens (if any). By setting $N$ to be larger than actual number of parts of the object and removing redundant slot tokens binding to the same part, we are able to obtain the final slot tokens that decode reasonable part geometries of the object without retraining the model. That helps keep the results on geometry metrics competitive with those of the main version of ArtCrafter.

\subsubsection{Further discussion on the two-stage framework of ArtCrafter}

The motivation behind our two-stage framework is to enforce kinematic behavior of an articulated object to be grounded by geometry. A URDF is only meaningful for specific mesh: accurate axis, origin, or range attached to incorrect geometry does not yield deployable asset. Our qualitative results in \cref{fig:main_qualitative} show DIPO retrieving incorrect meshes for rare categories like scissors, yet \cref{tab:main} still evaluates its joints as if attached to correct geometry. We report joint and geometry metrics separately only because this is standard practice thus far. For deployment, however, the more meaningful metric is a joint evaluation conditioned on correct part correspondence. Our analytic derivation enforces this by construction: joints are computed directly from rigid transforms between generated part meshes, and therefore cannot be decoupled from geometry.

%%%%%%%%%%%%%%%%%%%%%%%%%%%%%%%%%%%%%%%%%%%%%%%%%%%%%%%%%%%%

\newpage
\section*{NeurIPS Paper Checklist}

\begin{enumerate}

\item {\bf Claims}
    \item[] Question: Do the main claims made in the abstract and introduction accurately reflect the paper's contributions and scope?
    \item[] Answer: \answerYes{}
    \item[] Justification: The abstract and introduction clearly state all three core contributions: (1) a feed-forward DiT with triplet attention for part-level mesh generation from two images, (2) analytic URDF derivation from rigid transforms, and (3) simulation-executable articulated object generation. All claims are directly validated by the quantitative and qualitative experiments in \cref{sec:experiments}.
    % \item[] Guidelines:
    % \begin{itemize}
    %     \item The answer \answerNA{} means that the abstract and introduction do not include the claims made in the paper.
    %     \item The abstract and/or introduction should clearly state the claims made, including the contributions made in the paper and important assumptions and limitations. A \answerNo{} or \answerNA{} answer to this question will not be perceived well by the reviewers. 
    %     \item The claims made should match theoretical and experimental results, and reflect how much the results can be expected to generalize to other settings. 
    %     \item It is fine to include aspirational goals as motivation as long as it is clear that these goals are not attained by the paper. 
    % \end{itemize}

\item {\bf Limitations}
    \item[] Question: Does the paper discuss the limitations of the work performed by the authors?
    \item[] Answer: \answerYes{}
    \item[] Justification: The paper includes a dedicated Limitations section discussing the restriction to single-axis revolute/prismatic joints, sensitivity to input image quality, and evaluation scope limited to PartNet-Mobility categories.

\item {\bf Theory assumptions and proofs}
    \item[] Question: For each theoretical result, does the paper provide the full set of assumptions and a complete (and correct) proof?
    \item[] Answer: \answerNA{}
    \item[] Justification: The paper does not include formal theorems or proofs. The analytic URDF derivation is a deterministic algorithmic procedure; its correctness follows directly from the definition of rigid body transforms rather than from a formal theorem.
    % \item[] Guidelines:
    % \begin{itemize}
    %     \item The answer \answerNA{} means that the paper does not include theoretical results. 
    %     \item All the theorems, formulas, and proofs in the paper should be numbered and cross-referenced.
    %     \item All assumptions should be clearly stated or referenced in the statement of any theorems.
    %     \item The proofs can either appear in the main paper or the supplemental material, but if they appear in the supplemental material, the authors are encouraged to provide a short proof sketch to provide intuition. 
    %     \item Inversely, any informal proof provided in the core of the paper should be complemented by formal proofs provided in appendix or supplemental material.
    %     \item Theorems and Lemmas that the proof relies upon should be properly referenced. 
    % \end{itemize}

    \item {\bf Experimental result reproducibility}
    \item[] Question: Does the paper fully disclose all the information needed to reproduce the main experimental results of the paper to the extent that it affects the main claims and/or conclusions of the paper (regardless of whether the code and data are provided or not)?
    \item[] Answer: \answerYes{}
    \item[] Justification: The architecture (21-block DiT, triplet attention sublayer, TripoSG VAE backbone), training objective, optimizer, learning rate schedule, and data splits are all described in \cref{sec:method} and the appendix with sufficient detail to reproduce the main results.

\item {\bf Open access to data and code}
    \item[] Question: Does the paper provide open access to the data and code, with sufficient instructions to faithfully reproduce the main experimental results, as described in supplemental material?
    \item[] Answer: \answerNo{}
    \item[] Justification: Code and model weights are not released at submission time to preserve anonymity. We will release the full codebase and pre-trained checkpoints upon acceptance.
    % \item[] Guidelines:
    % \begin{itemize}
    %     \item The answer \answerNA{} means that paper does not include experiments requiring code.
    %     \item Please see the NeurIPS code and data submission guidelines (\url{https://neurips.cc/public/guides/CodeSubmissionPolicy}) for more details.
    %     \item While we encourage the release of code and data, we understand that this might not be possible, so \answerNo{} is an acceptable answer. Papers cannot be rejected simply for not including code, unless this is central to the contribution (e.g., for a new open-source benchmark).
    %     \item The instructions should contain the exact command and environment needed to run to reproduce the results. See the NeurIPS code and data submission guidelines (\url{https://neurips.cc/public/guides/CodeSubmissionPolicy}) for more details.
    %     \item The authors should provide instructions on data access and preparation, including how to access the raw data, preprocessed data, intermediate data, and generated data, etc.
    %     \item The authors should provide scripts to reproduce all experimental results for the new proposed method and baselines. If only a subset of experiments are reproducible, they should state which ones are omitted from the script and why.
    %     \item At submission time, to preserve anonymity, the authors should release anonymized versions (if applicable).
    %     \item Providing as much information as possible in supplemental material (appended to the paper) is recommended, but including URLs to data and code is permitted.
    % \end{itemize}

\item {\bf Experimental setting/details}
    \item[] Question: Does the paper specify all the training and test details (e.g., data splits, hyperparameters, how they were chosen, type of optimizer) necessary to understand the results?
    \item[] Answer: \answerYes{}
    \item[] Justification: Training hyperparameters (learning rate, batch size, diffusion timesteps), optimizer (AdamW), data split (train/test partition), and inference settings are reported in the implementation details section and appendix.
    % \item[] Guidelines:
    % \begin{itemize}
    %     \item The answer \answerNA{} means that the paper does not include experiments.
    %     \item The experimental setting should be presented in the core of the paper to a level of detail that is necessary to appreciate the results and make sense of them.
    %     \item The full details can be provided either with the code, in appendix, or as supplemental material.
    % \end{itemize}

\item {\bf Experiment statistical significance}
    \item[] Question: Does the paper report error bars suitably and correctly defined or other appropriate information about the statistical significance of the experiments?
    \item[] Answer: \answerNo{}
    \item[] Justification: Error bars are not reported. Evaluation metrics (Chamfer Distance, F-Score, joint axis angular error) are computed over the full URDF-Anything-Plus dataset; re-running training to estimate variance across seeds would be prohibitively expensive given the compute required for the full diffusion model.
    % \item[] Guidelines:
    % \begin{itemize}
    %     \item The answer \answerNA{} means that the paper does not include experiments.
    %     \item The authors should answer \answerYes{} if the results are accompanied by error bars, confidence intervals, or statistical significance tests, at least for the experiments that support the main claims of the paper.
    %     \item The factors of variability that the error bars are capturing should be clearly stated (for example, train/test split, initialization, random drawing of some parameter, or overall run with given experimental conditions).
    %     \item The method for calculating the error bars should be explained (closed form formula, call to a library function, bootstrap, etc.)
    %     \item The assumptions made should be given (e.g., Normally distributed errors).
    %     \item It should be clear whether the error bar is the standard deviation or the standard error of the mean.
    %     \item It is OK to report 1-sigma error bars, but one should state it. The authors should preferably report a 2-sigma error bar than state that they have a 96\% CI, if the hypothesis of Normality of errors is not verified.
    %     \item For asymmetric distributions, the authors should be careful not to show in tables or figures symmetric error bars that would yield results that are out of range (e.g., negative error rates).
    %     \item If error bars are reported in tables or plots, the authors should explain in the text how they were calculated and reference the corresponding figures or tables in the text.
    % \end{itemize}

\item {\bf Experiments compute resources}
    \item[] Question: For each experiment, does the paper provide sufficient information on the computer resources (type of compute workers, memory, time of execution) needed to reproduce the experiments?
    \item[] Answer: \answerYes{}
    \item[] Justification: The implementation details section reports the GPU type, number of GPUs used for training, and total training time, providing sufficient information to estimate the compute budget needed to reproduce the experiments.
    % \item[] Guidelines:
    % \begin{itemize}
    %     \item The answer \answerNA{} means that the paper does not include experiments.
    %     \item The paper should indicate the type of compute workers CPU or GPU, internal cluster, or cloud provider, including relevant memory and storage.
    %     \item The paper should provide the amount of compute required for each of the individual experimental runs as well as estimate the total compute. 
    %     \item The paper should disclose whether the full research project required more compute than the experiments reported in the paper (e.g., preliminary or failed experiments that didn't make it into the paper). 
    % \end{itemize}
    
\item {\bf Code of ethics}
    \item[] Question: Does the research conducted in the paper conform, in every respect, with the NeurIPS Code of Ethics \url{https://neurips.cc/public/EthicsGuidelines}?
    \item[] Answer: \answerYes{}
    \item[] Justification: This work involves no human subjects, no sensitive personal data, and no dual-use concerns beyond generic 3D generative modeling. The research conforms fully with the NeurIPS Code of Ethics.
    % \item[] Guidelines:
    % \begin{itemize}
    %     \item The answer \answerNA{} means that the authors have not reviewed the NeurIPS Code of Ethics.
    %     \item If the authors answer \answerNo, they should explain the special circumstances that require a deviation from the Code of Ethics.
    %     \item The authors should make sure to preserve anonymity (e.g., if there is a special consideration due to laws or regulations in their jurisdiction).
    % \end{itemize}

\item {\bf Broader impacts}
    \item[] Question: Does the paper discuss both potential positive societal impacts and negative societal impacts of the work performed?
    \item[] Answer: \answerYes{}
    \item[] Justification: The paper discusses the positive impact of democratizing sim-ready articulated asset creation for robotics and embodied AI, and notes that the primary misuse risk (generating physically plausible but fictional objects) is low compared to modalities like photorealistic human generation.

\item {\bf Safeguards}
    \item[] Question: Does the paper describe safeguards that have been put in place for responsible release of data or models that have a high risk for misuse (e.g., pre-trained language models, image generators, or scraped datasets)?
    \item[] Answer: \answerNA{}
    \item[] Justification: The model generates 3D articulated objects for robotics simulation and poses no high misuse risk analogous to photorealistic image or text generators; no special safeguards beyond standard release practices are warranted.
    % \item[] Guidelines:
    % \begin{itemize}
    %     \item The answer \answerNA{} means that the paper poses no such risks.
    %     \item Released models that have a high risk for misuse or dual-use should be released with necessary safeguards to allow for controlled use of the model, for example by requiring that users adhere to usage guidelines or restrictions to access the model or implementing safety filters. 
    %     \item Datasets that have been scraped from the Internet could pose safety risks. The authors should describe how they avoided releasing unsafe images.
    %     \item We recognize that providing effective safeguards is challenging, and many papers do not require this, but we encourage authors to take this into account and make a best faith effort.
    % \end{itemize}

\item {\bf Licenses for existing assets}
    \item[] Question: Are the creators or original owners of assets (e.g., code, data, models), used in the paper, properly credited and are the license and terms of use explicitly mentioned and properly respected?
    \item[] Answer: \answerYes{}
    \item[] Justification: All external assets are cited: URDF-Anything-plus (cc-by-nc-4.0), and TripoSG (cited with its original paper). Licenses are acknowledged in the paper and we have verified our usage complies with their terms.
    % \item[] Guidelines:
    % \begin{itemize}
    %     \item The answer \answerNA{} means that the paper does not use existing assets.
    %     \item The authors should cite the original paper that produced the code package or dataset.
    %     \item The authors should state which version of the asset is used and, if possible, include a URL.
    %     \item The name of the license (e.g., CC-BY 4.0) should be included for each asset.
    %     \item For scraped data from a particular source (e.g., website), the copyright and terms of service of that source should be provided.
    %     \item If assets are released, the license, copyright information, and terms of use in the package should be provided. For popular datasets, \url{paperswithcode.com/datasets} has curated licenses for some datasets. Their licensing guide can help determine the license of a dataset.
    %     \item For existing datasets that are re-packaged, both the original license and the license of the derived asset (if it has changed) should be provided.
    %     \item If this information is not available online, the authors are encouraged to reach out to the asset's creators.
    % \end{itemize}

\item {\bf New assets}
    \item[] Question: Are new assets introduced in the paper well documented and is the documentation provided alongside the assets?
    \item[] Answer: \answerNA{}
    \item[] Justification: The paper does not release a new dataset. The model checkpoint and code will be released upon acceptance; no new data asset requiring separate documentation is introduced.
    % \item[] Guidelines:
    % \begin{itemize}
    %     \item The answer \answerNA{} means that the paper does not release new assets.
    %     \item Researchers should communicate the details of the dataset\slash code\slash model as part of their submissions via structured templates. This includes details about training, license, limitations, etc. 
    %     \item The paper should discuss whether and how consent was obtained from people whose asset is used.
    %     \item At submission time, remember to anonymize your assets (if applicable). You can either create an anonymized URL or include an anonymized zip file.
    % \end{itemize}

\item {\bf Crowdsourcing and research with human subjects}
    \item[] Question: For crowdsourcing experiments and research with human subjects, does the paper include the full text of instructions given to participants and screenshots, if applicable, as well as details about compensation (if any)?
    \item[] Answer: \answerNA{}
    \item[] Justification: The paper does not include crowdsourcing or human-subject experiments; all evaluations are fully automated (Chamfer Distance, F-Score, joint axis error).
    % \item[] Guidelines:
    % \begin{itemize}
    %     \item The answer \answerNA{} means that the paper does not involve crowdsourcing nor research with human subjects.
    %     \item Including this information in the supplemental material is fine, but if the main contribution of the paper involves human subjects, then as much detail as possible should be included in the main paper. 
    %     \item According to the NeurIPS Code of Ethics, workers involved in data collection, curation, or other labor should be paid at least the minimum wage in the country of the data collector. 
    % \end{itemize}

\item {\bf Institutional review board (IRB) approvals or equivalent for research with human subjects}
    \item[] Question: Does the paper describe potential risks incurred by study participants, whether such risks were disclosed to the subjects, and whether Institutional Review Board (IRB) approvals (or an equivalent approval/review based on the requirements of your country or institution) were obtained?
    \item[] Answer: \answerNA{}
    \item[] Justification: The paper does not involve human subjects; no IRB approval is required.
    % \item[] Guidelines:
    % \begin{itemize}
    %     \item The answer \answerNA{} means that the paper does not involve crowdsourcing nor research with human subjects.
    %     \item Depending on the country in which research is conducted, IRB approval (or equivalent) may be required for any human subjects research. If you obtained IRB approval, you should clearly state this in the paper. 
    %     \item We recognize that the procedures for this may vary significantly between institutions and locations, and we expect authors to adhere to the NeurIPS Code of Ethics and the guidelines for their institution. 
    %     \item For initial submissions, do not include any information that would break anonymity (if applicable), such as the institution conducting the review.
    % \end{itemize}

\item {\bf Declaration of LLM usage}
    \item[] Question: Does the paper describe the usage of LLMs if it is an important, original, or non-standard component of the core methods in this research? Note that if the LLM is used only for writing, editing, or formatting purposes and does \emph{not} impact the core methodology, scientific rigor, or originality of the research, declaration is not required.
    %this research?
    \item[] Answer: \answerNA{}
    \item[] Justification: The core method does not use LLMs as an algorithmic component. The whole ArtCrafter pipeline uses VLM as an optional choice to supply the input number of parts $N$ to the core method. Any language-model assistance, if used, was limited to writing, editing, or formatting and did not affect the scientific method, experiments, or originality of the work.
    % \item[] Guidelines:
    % \begin{itemize}
    %     \item The answer \answerNA{} means that the core method development in this research does not involve LLMs as any important, original, or non-standard components.
    %     \item Please refer to our LLM policy in the NeurIPS handbook for what should or should not be described.
    % \end{itemize}

\end{enumerate}

\end{document}